\documentclass[times,twocolumn,final,authoryear]{elsarticle}

\usepackage{ycviu}
\usepackage{framed}

\usepackage{latexsym}

\definecolor{newcolor}{rgb}{.8,.349,.1}

\usepackage{graphicx}
\usepackage{times}
\usepackage[utf8]{inputenc} % allow utf-8 input
\usepackage[T1]{fontenc}    % use 8-bit T1 fonts
\usepackage{url}            % simple URL typesetting
\usepackage{booktabs}       % professional-quality tables
\usepackage{amsfonts}       % blackboard math symbols
\usepackage{nicefrac}       % compact symbols for 1/2, etc.
\usepackage{microtype}      % microtypography        % colors
\usepackage{epsfig}
\usepackage{float}
\usepackage{multicol}
\usepackage{multirow}
\usepackage{amssymb}
\usepackage{amsmath}
\usepackage{wrapfig}
\usepackage{xspace}
\usepackage{color}
\usepackage{colortbl}
\usepackage[dvipsnames, svgnames, x11names, table]{xcolor}
\usepackage[misc]{ifsym}
\usepackage{makecell}
\usepackage{soul}
\usepackage{bm}
\usepackage{wrapfig}
\usepackage{subcaption, overpic, textpos}
\usepackage{ulem}
\usepackage{paralist}

\usepackage[breaklinks=true,colorlinks,citecolor=blue,urlcolor=blue,linkcolor=blue,bookmarks=false]{hyperref}

\usepackage{adjustbox}
\usepackage{pifont}
\usepackage{array}
\usepackage{enumitem}
\newcolumntype{C}[1]{>{\centering\arraybackslash}p{#1}} % 表格列宽

\def \pzo {\phantom{0}}

\newcommand{\cmark}{\ding{52}\xspace}%
\newcommand{\xmark}{\ding{56}\xspace}%
\newcommand{\xmarkg}{\textcolor{lightgray}{\ding{56}}\xspace}%
\newcommand{\pmark}{\ding{58}\xspace}%
\newcommand{\omark}{$\bigcirc$\xspace}%

\definecolor{lightgray}{rgb}{0.8, 0.8, 0.8}
\definecolor{lgray}{rgb}{0.66, 0.66, 0.66}
\definecolor{whit_tab}{RGB}{255, 255, 255}
\definecolor{gray_tab}{RGB}{235, 235, 235}
\definecolor{oran_tab}{RGB}{248, 230, 218}
\definecolor{blue_tab}{RGB}{200, 227, 245}
\definecolor{lblu_tab}{RGB}{225, 235, 246}
\definecolor{orange_vitad}{RGB}{222, 131, 68}
\definecolor{blue_vitad}{RGB}{106, 153, 208}
\newcommand{\red}{\textcolor[RGB]{255, 36, 24}}

\newcommand{\bluetab}{\textcolor[RGB]{225, 235, 246}}

\definecolor{tab_others}{RGB}{235, 235, 235}
\definecolor{tab_ours}{RGB}{225, 235, 246}

\usepackage[ruled,vlined,linesnumbered]{algorithm2e}
\usepackage[capitalize]{cleveref}
\crefname{section}{Sec.}{Secs.}
\Crefname{section}{Section}{Sections}
\Crefname{table}{Table}{Tables}
\crefname{table}{Tab.}{Tabs.}

\newlength\savewidth

\renewcommand{\paragraph}[1]{\vspace{1.25mm}\noindent\textbf{#1}}

\newcommand{\ie}{i.e}
\newcommand{\eg}{e.g}
\newcommand{\Eg}{E.g}

\def\onedot{.\xspace}
\def\eg{\textit{e.g}\onedot} 
\def\Eg{\textit{E.g}\onedot}
\def\ie{\textit{i.e}\onedot}

\def\cf{\textit{c.f}\onedot}

\def\aka{\textit{a.k.a}\onedot}

\newcommand\tb[1]{\textcolor{blue}{#1}}
\newcommand\tr[1]{\textcolor{red}{#1}}

\newcommand\rub[1]{\red{${+#1\uparrow}$}}

\usepackage{tablefootnote}
\usepackage{diagbox}

\begin{document}

\begin{frontmatter}

\title{Exploring Plain ViT Features for Multi-class Unsupervised Visual Anomaly Detection}

\author[1]{Jiangning \snm{Zhang}} 
\cortext[cor1]{First two authors contribute equally.}
\author[2]{Xuhai \snm{Chen}}
\author[1]{Yabiao \snm{Wang}}
\author[1]{Chengjie \snm{Wang}}
\author[2]{Yong \snm{Liu}\corref{cor1}}
\cortext[cor2]{Corresponding author.}
% Tel.: +86-138-0571-9977;}
\ead{yongliu@iipc.zju.edu.cn}
\author[3]{Xiangtai \snm{Li}}
\author[4]{Ming-Hsuan \snm{Yang}}
\author[3]{Dacheng \snm{Tao}}

\address[1]{Youtu Lab, Tencent, Shanghai, China}
\address[2]{Zhejiang University, Hangzhou, China}
\address[3]{Nanyang Technological University, Singapore}
\address[4]{Department of Computer Science and Engineering at the University of California, Merced, US}

\received{1 May 2013}
\finalform{10 May 2013}
\accepted{13 May 2013}
\availableonline{15 May 2013}
\communicated{S. Sarkar}

\begin{abstract}
% This work studies a challenging and practical problem, termed multi-class unsupervised anomaly detection (MUAD), which only requires normal images for training while simultaneously testing both normal and anomaly images for multiple classes. Existing reconstruction-based methods typically adopt pyramidal networks as encoders and decoders to obtain multi-resolution features, accompanied by elaborating sub-modules with heavier handcraft engineering designs. In contrast, a plain Vision Transformer (ViT) showcasing a more straightforward architecture has proven effective in multiple domains, including detection and segmentation tasks. It is simpler, more effective, and elegant. Following this spirit, we explore plain ViT features for MUAD. We first abstract a Meta-AD concept by inducing current reconstruction-based methods. Then, we instantiate a novel ViT-based ViTAD structure, effectively designed step by step from global and local perspectives. In addition, this paper reveals several interesting findings for further exploration. Finally, we benchmark various approaches comprehensively and fairly on eight metrics. Based on a naive training recipe with only an MSE loss, ViTAD achieves state-of-the-art results and efficiency on MVTec AD, VisA, and Uni-Medical datasets, obtaining 85.4 mAD that surpasses UniAD by +3.0, while only requiring 1.1 hours and 2.3G GPU memory to complete model training by one V100 on the MVTec AD dataset. 

This work studies a challenging and practical issue known as multi-class unsupervised anomaly detection (MUAD). 
This problem requires only normal images for training while simultaneously testing both normal and anomaly images across multiple classes. 
Existing reconstruction-based methods typically adopt pyramidal networks as encoders and decoders to obtain multi-resolution features, often involving complex sub-modules with extensive handcraft engineering. 
In contrast, a plain Vision Transformer (ViT) showcasing a more straightforward architecture has proven effective in multiple domains, including detection and segmentation tasks. It is simpler, more effective, and elegant. Following this spirit, we explore the use of only plain ViT features for MUAD. We first abstract a Meta-AD concept by synthesizing current reconstruction-based methods. 
Subsequently, we instantiate a novel ViT-based ViTAD structure, designed incrementally from both global and local perspectives. This  model provide a strong baseline to facilitate future research. 
Additionally, this paper uncovers several intriguing findings for further investigation. Finally, we comprehensively and fairly benchmark various approaches using eight metrics. Utilizing a basic training regimen with only an MSE loss, ViTAD achieves state-of-the-art results and efficiency on MVTec AD, VisA, and Uni-Medical datasets. \Eg, achieving 85.4 mAD that surpasses UniAD by +3.0 for the MVTec AD dataset, and it requires only 1.1 hours and 2.3G GPU memory to complete model training on a single V100 that can serve as a strong baseline to facilitate the development of future research. Full code is available at \url{https://zhangzjn.github.io/projects/ViTAD/}.

\end{abstract}


\end{frontmatter}

%\linenumbers

\section{Introduction} \label{section:intro}
Visual Anomaly Detection (AD) aims to identify unusual or unexpected patterns within images that deviate significantly from the norm images. 
This technique helps prevent potential risks, improve safety, or enhance system performance, relying on its ability to reveal critical information across various domains. 
Thus, it has been widely used in visual inspection~\citep{mvtec,simplenet}, medical image lesion detection~\citep{bmad}, and video surveillance~\citep{video1}, to name a few. 

As the unsupervised anomaly detection approach does not require high labeling costs, it has received increasing attention in recent years~\citep{survey_ad}.
Existing methods are generally developed based on a Single-class Unsupervised Anomaly Detection (SUAD) setting~\citep{draem,rd,simplenet}, where each class requires a separate model for training that significantly increases the training and storage costs of the model. 
To alleviate the above problem, recent UniAD~\citep{uniad} proposes the multi-class setting for the first time, but significant opportunities still exist for enhancing performance and reducing training costs. 
This work tackles this more challenging and practical Multi-class Unsupervised Anomaly Detection (MUAD) task.

\begin{figure*}[tp]
    \centering
    \includegraphics[width=.95\linewidth]{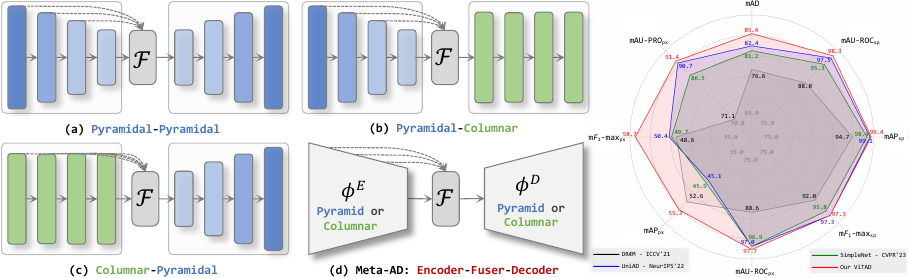}
    \caption{Left: (a-c) display general reconstruction-based AD frameworks. (d) shows a Meta-AD framework that consists of image Encoder \bm{$\phi^{E}$}, Fuser \textit{\bm{$\mathcal{F}$}}, and Decoder \bm{$\phi^{D}$}. The dashed line indicates that the feature may be used by the Fuser \bm{$\mathcal{F}$}. 
    Right: Comprehensive quantitative comparison with popular methods by eight metrics on MVTec AD dataset~\citep{mvtec} (see \cref{section:setup} and \cref{exp:sotas}).
    }
    \label{fig:motivation}
\end{figure*}
 
Unsupervised anomaly detection methods in the literature can be broadly categorized as follows. 
i) Augmentation-based methods~\citep{cutpaste,simplenet,uniad} enhance performance by introducing fabricated anomaly information. 
ii) Embedding-based methods~\citep{uninformedstudents,patchcore} map normal features to a compact space and identify anomalies by feature comparison. 
iii) Reconstruction-based methods generally follow an encoder-decoder framework to reconstruct input images or features, and simple reconstruction errors serve as the anomaly map during inference. 
Thanks to the effectiveness, interpretability, and scalability, numerous subsequent researchers~\citep{rd,rrd,diad} follow this framework. 
According to whether the encoder-decoder structure is pyramidal (\eg, multi-resolution ResNet~\citep{resnet}) or columnar (\eg, single-resolution ViT~\citep{vit}), existing methods can be categorized as 
fully pyramidal structures~\citep{dae,ganomaly,rd} (\cref{fig:motivation}(a)), 
pyramidal encoder with a dynamic ViT~\citep{adtr,uniad} (\cref{fig:motivation}(b)), 
and plain ViT-based encoder with a pyramidal decoder~\citep{vt_adl}(\cref{fig:motivation}(c)). 
However, over-reliance on pyramidal structure~\citep{rd} lacks long-distance perception in early stages, leading to wrong results casually (see \cref{fig:cable}). 
Thus, some works introduce ViT partially to enhance global modeling capabilities, but their results are still unsatisfactory, and the training cost is expensive~\citep{uniad,vt_adl}.
This motivates us to explore a more powerful and efficient MUAD model while considering global interactions. 

Starting by abstracting current methods to a new Meta-AD concept (\cref{fig:motivation}(d)) that consists of: 
i) a feature \textit{Encoder} to map the input to the latent space while compressing the spatial dimensions; 
ii) a feature \textit{Fuser} to fuse multi-stage feature maps and generate more compact features as the decoder input; 
iii) a feature \textit{Decoder} to reconstruct the original image or features by constraining the reconstruction loss. 
Existing methods~\citep{survey_ad,rd,uniad} typically introduce pyramidal networks as encoders or decoders to obtain multi-resolution features for more accurate anomaly locations. 
Nevertheless, plain ViT serves as a visual foundation model that has been effective for possessing long-distance modeling capabilities in many downstream tasks, \eg, object detection~\citep{vitdet}, semantic segmentation~\citep{segvit}, and human pose estimation~\citep{vitpose}. 
This motivates us to explore the feasibility of exploring only plain ViT features for the anomaly detection field.

Within the Meta-AD framework, we propose an efficient ViT-based AD model that only includes global modeling without a pyramidal encoder and decoder. 
Notably, we adopt a 12-layer ViT-S evenly divided into four stages to supply multi-depth features like ViTDet~\citep{vitdet}. 
However, directly applying the naive ViT to Meta-AD leads to extremely poor anomaly classification and localization results (see \cref{exp:global}).
This leads to subsequent explorations to find an effective ViT structure for the MUAD task. 
Adhering to Occam's Razor principle, we improve plain ViT from global and local perspectives to explore its anomaly detection capability without resorting to complex modules or training strategies. 
%
%MH: the way you structure this paper is not idea as it reads like you have the ideas from empirical results (removing X, Y, Z from existing methods). There is nothing wrong with that but it reads as you do not have strong orginiality. 
\textit{From a global perspective}, we empirically design three structural improvements that notably boost the model performance: 
i) A plain ViT with the same resolution shortens the information flow path, leading to potential information leakage. Removing skip connections of multi-depth features for RD-like~\citep{rd} fusion manner improves performance. 
ii) The last three stages can provide rich multi-depth information, and using them for loss constraints and anomaly map calculations enables a higher anomaly localization capability of the model, which differs from the default usage of the first three stages in current works~\citep{rd,uniad}. 
iii) Due to the gap between AD datasets and ImageNet~\citep{imagenet}, weights obtained from more generic and category-independent unsupervised training~\citep{dino} yield better results. 
\textit{From a local perspective,} we explore four factors to improve the model performance further: 
i) whether the output for the encoder goes through the final batch normalization. 
ii) whether the feature fuser uses linear feature transformation. 
iii) whether the class token is inherited. 
iv) whether position embeddings are used in the decoder. 

Extensive experimental results reveal the effects of the proposed modules with three main interesting findings in \cref{exp:ablation}.
i) Pyramidal structure for encoder/decoder is \textbf{not} necessary for AD models. The simple plain ViT can yield impressive state-of-the-art (SoTA) results (\cref{exp:sotas}). 
ii) The pre-training weights and model scale of the encoder significantly impact the results, and their performance in classification tasks does not consistently correlate with AD results (\cref{exp:global}). 
iii) A heavy feature fuser is not necessary that a superficial linear layer suffices, which contradicts the design conclusion of previous works~\citep{HaloAE,rd} (\cref{exp:local}). 

We make the following three contributions in this work:
% \begin{compactenum}
\begin{compactitem}
\item We present a novel ViTAD model inspired by Meta-AD, which explores pure plain ViT as the fundamental structure. 
Specifically, ViTAD is effectively designed from global and local perspectives (\cref{section:vitad}). 
In addition, we provide in-depth discussions and present some interesting findings.
\item We replicate and benchmark state-of-the-art methods for fair comparisons of the MUAD task. 
We propose using eight metrics to comprehensively evaluate different approaches (\cref{section:setup}). 
\item Extensive experimental evaluations with state-of-the-art methods on MVTec AD, VisA, and Uni-Medical datasets demonstrate the performance, efficiency, and robustness of our ViTAD. 
Furthermore, we explore and ablate factors that may affect model performance (\cref{exp:ablation}). 
\end{compactitem}

\section{Related Work} \label{section:related}
\noindent
%\textbf{Visual Anomaly Detection} is dedicated to classifying and segmenting abnormal images from normal ones, typically divided into supervised and unsupervised categories~\citep{survey_ad}. 
\textbf{Visual Anomaly Detection.} Numerous methods have been developed to identify abnormal image regions from normal ones, based on supervised, semi-supervised, and unsupervised settings~\citep{survey_ad}. 
Supervised and semi-supervised approaches generally use synthetic data~\citep{draem} or few-shot anomaly samples~\citep{graphcore,winclip,aprilgan,clipad,saa,gpt-4v-ad,anomalydiffusion}, which can be seen as a particular case of a binary classification task. 
Unsupervised AD methods rely only on normal training data to distinguish abnormal regions during testing, which can usually be broadly categorized as: 
i) Augmentation-based methods~\citep{cutpaste,draem,uniad,simplenet} typically involve synthesizing abnormal regions on normal images or adding anomalous information to normal features to construct pseudo supervisory signals for better one-class classification. 
ii) Embedding-based methods map normal features to a compact space, distancing them from abnormal parts to ensure discriminative capabilities. 
Existing approaches are based on distribution models~\citep{wan2022position,pyramidflow}, teacher student models~\citep{cao2022informative,destseg,uninformedstudents}, and memory banks~\citep{memkd,patchcore}. 
iii) Reconstruction-based methods generally consist of an encoder and a decoder, while some approaches incorporate an additional transformation module. 
Anomaly localization is achieved by measuring the discrepancy between the input and reconstructed data, and current methods can be roughly categorized into two taxonomies based on the data being reconstructed, \ie, RGB images and deep features.
Image-level methods~\citep{InTra,ocrgan,madan2023self} essentially are developed based on reconstruction of normal RGB images~\citep{mei2018unsupervised,zavrtanik2022dsr,diad}. 
A few approaches~\citep{anogan,skipganomaly} introduce deep generative adversarial networks to learn normal manifold, while subsequent studies incorporate anomaly data augmentation strategies to enhance pseudo-supervision of the model, such as random noise generation~\citep{deng2023noise}, random mask generation~\citep{zavrtanik2021reconstruction}, cut paste operation~\citep{ocrgan}, and DRAEM-like predefined anomaly data construction schemes~\citep{draem}. 
Feature-level methods reconstruct more expressive deep features that generally achieve better performance~\citep{adtr,rrd}. 
UniAD~\citep{uniad} corroborates the pivotal function of query embedding in circumventing shortcut feature-level distribution and further introduces a layer-wise query decoder to model feature-level distribution. 
On the other hand, RD~\citep{rd} introduces a novel ``reverse distillation" paradigm to reconstruct multi-resolution features, which can be viewed as the extension of multi-resolution representations. 
While augmentation-/embedding-based methods achieve satisfactory results, incorporating additional anomaly-related operations leads to complex and high-dimensional embeddings~\citep{liu2023diversity} or noise-sensitive models~\citep{noisy5}.
Capitalizing on the simplicity and effectiveness of the reconstruction scheme, we propose an effective and efficient ViT-based method for unsupervised anomaly detection.

\vspace{1mm}
\noindent
\textbf{Multi-class Unsupervised Anomaly Detection.}
Most existing methods require individual models for training each category~\citep{patchcore,draem,rd,ocrgan,rrd,simplenet}, \aka, Single-class Unsupervised Anomaly Detection (SUAD), which is unsuitable for practical applications. 
In contrast, UniAD~\citep{uniad} employs one unified model to cover multiple categories, \aka, Multi-class Unsupervised Anomaly Detection (MUAD). 
A few MUAD methods~\citep{omnial,uniad} have since then been developed. 
Our method is also formulated within this challenging but practical setting.

\vspace{1mm}
\noindent
\textbf{Pyramidal Architecture for Anomaly Detection.}
%AD field has a consensus among researchers that multi-resolution (pyramidal) features are necessary to model accurate anomaly locations. 
Numerous methods have demonstrated that multi-resolution (pyramidal) features are effective for anomaly detection~\citep{ae_ad,ocrgan,patchcore,rd,uniad,simplenet}. 
%
%MH: what are certain defects? Be clear  // Thanks, modified!
As these approaches generally include a heavy backbone and local model with a smaller receptive field, large-scale and long-distance defects may not be well detected~\citep{patchcore,rd,simplenet}. 
%
%MH: what is "global information interaction"? If you mean global depdence of visual features, then make it clear.  // Thanks, modified!
Recently, a few methods have used a dynamic ViT~\citep{vit} to model the global dependence of visual features to improve performance.
%
%MH: what do you need to use capital words for encoder and decoder? But what's wrong with encoder and decoder? You need to explain it clearly?  // Lower-case initials make more sense, modified!
However, they retain the pyramidal network in the encoder~\citep{adtr,uniad,destseg} or decoder~\citep{vt_adl,destseg}. 
%
%MH: stop saying for the first time. What is the big deal? and why plain and columnat ViT help you resovle the above mentioned issues?  // Thanks, modified!
Unlike these methods, this inspires us to explore solely using pure plain and columnar ViT for the MUAD task, expecting to leverage its long-distance dependence and strong modeling ability verified in other fields~\citep{vit,sam,vitdet}.

\vspace{1mm}
\noindent
\textbf{Plain Vision Transformer.}
%MH: stop saying who does first unless you are absolutely sure. In fact, why do you need to say that anyway? are they closely related to your work anyway? This is why you have such a long and unnecessary reference list. You should reduce the list to at most 50 to 60 papers. It is not a survey paper
%R: Thanks, I will modify this problem!
%Since Vision Transformer (ViT)~\citep{vit} Transformer~\citep{attention} structure into visual classification successfully, massive improvements have been subsequently developed~\citep{deit,pvt,swin,swinv2,eat,eatformer,emo}. 
%
Benefiting from global dynamic modeling capabilities, columnar plain ViT~\citep{vit} offers more excellent usability and practical values compared to the more complex pyramidal structures. 
%
%MH: what is the relevance of the following sentences to your work?
%Recently, researchers have been simplifying their approaches and attempting to employ plain ViT to tackle various downstream tasks, \ie, object detection~\citep{vitdet}, semantic segmentation~\citep{sam,li2023transformer,wu2023towards,li2023sfnet}, in-context visual learning~\citep{seggpt}, human pose estimation~\citep{vitpose,vitposeplus}, \etc. 
%
%MH: useless but correct statement
%Plain ViT has achieved impressive results above dense prediction tasks, demonstrating its strong multi-scale and pixel-level feature representation abilities. 
%
Although anomaly detection also potentially benefits from this capability, columnar ViT is only exploited in the encoder~\citep{vt_adl} or decoder~\citep{adtr,uniad} of existing methods.  
% 
%MH: I have no idea what you mean by "inherent bias"  // Thank you, Modified!
This inspires us to break the mold, representing the first exploration of the potential and application value of plain ViT in the challenging MUAD task.
In addition, re-trained ViT features have significant effects on downstream tasks. 
%
%MH: I do not understand this statement
%R: Different ViT pre-training methods produce different extraction feature distributions, which will have different effects on AD tasks. I rewrite the sentence to avoid being misleading.
Except for supervisorily-trained on ImageNet~\citep{imagenet}, numerous unsupervised pre-training methods~\citep{moco,dino,dino2,clip,mae} endow ViT with different feature extraction distributions, potentially affecting the effectiveness of AD models. 
This paper explores the effects of different pre-training manners and shows that DINO-based features perform best for MUAD. 

\vspace{1mm}
\noindent
\textbf{Efficient Network Design for Anomaly Detection.}
%While model efficiency is important for practical applications, yet current methods are pursuing algorithmic accuracy while overlooking this aspect. 
Existing methods pay more attention to accuracy rather than model efficiency. 
For instance, DRAEM~\citep{draem} contains 97.4M parameters and requires 19.6 GPU hours under the condition of 300 epochs of training. 
It takes 1000 epochs for UniAD~\citep{uniad} to converge to satisfactory results, and RD~\citep{rd} requires a large parameter count of 80.6M due to its use of a pyramidal WideResNet-50 encoder/decoder. 
Recent EfficientAD~\citep{efficientad} proposes a lightweight feature extractor to reduce the model cost, but its performance is not satisfactory and needs further improvement for the MUAD setting.
%
%The excessive number of parameters and training time indicate that the model requires more storage and computational power, thus reducing its practical value. 
%
Thanks to the simplicity of the basic transformer block, the small-scale ViT has competitive running efficiency. 
In contrast to counterparts, we use ViT-S~\citep{vit} as the backbone, which is more compact and effective. 
It only requires training of 100 epochs with 1.1 GPU hours and performs favorably against state-of-the-art schemes.

\begin{figure}[tp]
    \centering
    \includegraphics[width=0.9\linewidth]{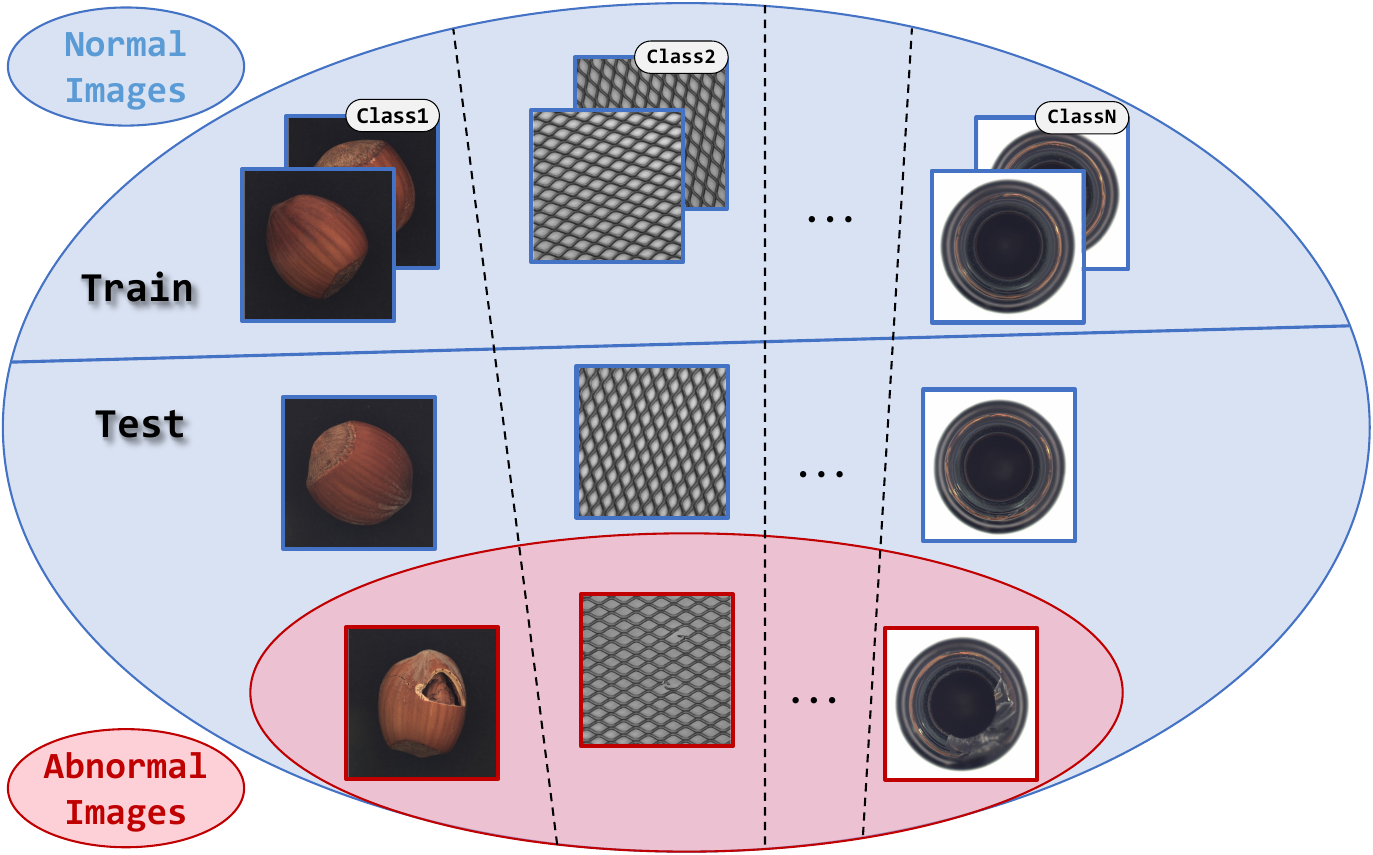}
    \caption{Diagram of Multi-class Unsupervised AD setting.}
    \label{fig:task}
\end{figure}

\section{Methodology: Abstract then Instantiate} \label{section:method}

\subsection{Task Definition of MUAD} \label{section:muad}
Similar to tasks such as object classification~\citep{imagenet} and detection~\citep{coco}, a more valuable anomaly detection setting should involve a model handling multiple categories simultaneously. 
%
%MH: the following statement for SUAD is not that important 
%R: Most methods follow general SUAD, so SUAD is introduced to incrementally introduce and highlight MUAD.
Given an AD dataset that contains $N$ classes $\boldsymbol{C} = \{ \textit{C}_1, \textit{C}_2, \cdots, \textit{C}_N\}$, the Single-class Unsupervised AD (SUAD) setting uses only one class $\textit{C}_{i}$ ($i=1, 2, \cdots, N$) that contains one-class set $\chi_{i} = \{(\chi_{i, normal}^{Train}), (\chi_{i, normal}^{Test}, \chi_{i, anomaly}^{Test})\}$, \ie, $\boldsymbol{C}_{\textit{Train}} = \boldsymbol{C}_{\textit{Test}} = \textit{C}_{i}$; 
while the Multi-class Unsupervised AD (MUAD) setting covers all classes $\boldsymbol{C}$ that contain all-class sets $\chi = \sum_{i=1}^{N}\chi_{i}$ in one unified model, \ie, $\boldsymbol{C}_{\textit{Train}} = \boldsymbol{C}_{\textit{Test}} = \boldsymbol{C}$. 
As shown in \cref{fig:task}, the normal images of all classes \textit{marked in \tb{\textbf{blue}}} are used for training simultaneously without any extra labeled samples, \ie, \textit{no anomalous and defective images marked in \tr{\textbf{red}}} are used.
On the other hand, normal and abnormal samples are used together for performance evaluation.  
This challenging task is described in UniAD~\citep{uniad} and adopted in recent works~\citep{sivt,omnial}. 
To avoid confusion, we use image-level detection and pixel-level location to refer to classification and segmentation as used in prior works.

\subsection{Formulation of Meta-AD} 
\label{section:metaad}
\begin{figure}[tp]
    \centering
    \includegraphics[width=1.0\linewidth]{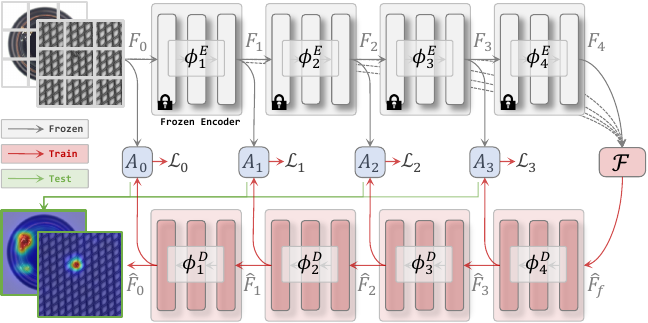}
    \caption{\textbf{Reconstruction-based Meta-AD paradigm,} which consists of a pretrained image encoder {\bm{$\phi^{E}$}} to obtain features at different depths from the patch embedding input, a feature fuser  {\textit{\bm{$\mathcal{F}$}}} to aggregate extracted multiple features, and a decoder  {\bm{$\phi^{D}$}} that has the same structure with the encoder to reconstruct multi-depth features. During the training phase, $\hat{F}_{i}$ is constrained by $F_{i}$ with loss function $\mathcal{L}_{i}$ to update  {\bm{$\phi^{D}$}}, while both $\hat{F}_{i}$ and $F_{i}$ are used to calculate anomaly map $A_{i}$ for inference.}
    \label{fig:metaad}
\end{figure}

As shown in \cref{fig:motivation}, we abstract existing reconstruction-based approaches by a meta framework that contains a feature \textit{Encoder}, a \textit{Fuser}, and a \textit{Decoder}.
\cref{fig:metaad} details the specific process of its structural treatment and symbols. 
% Also, we reuse the diagrams and symbols from \cref{fig:metaad} to illustrate the procedure. 

\vspace{1mm}
\noindent{\textbf{\textit{1)} Feature Encoder.}} This sub-module maps the input stem feature map $F_0$ to multi-scale deep features, which usually uses a frozen network trained on ImageNet~\citep{imagenet}.
The encoder \bm{$\phi^{E}$} consists of $N$ stages $\{{\bm{\phi^{E}_{1}}}, {\bm{\phi^{E}_{2}}}, $\dots$, {\bm{\phi^{E}_{N}}}\}$ with each containing a number of basic blocks
This module can be either a pyramidal structure with a decreasing resolution (\eg, ResNet~\citep{resnet} and EfficientNet~\citep{efficientnet}) or a columnar structure with a consistent resolution (\eg, ViT~\citep{vit}); 
For the local component, it can be a CNN~\citep{rd} or a transformer~\citep{utrad}.
Pyramidal structures~\citep{rd,uniad,utrad} are more commonly used for the strong ability to extract rich multi-scale features $F=\{F_{0},F_{1},\dots,F_{N}\}$. 
Denote $i$-th feature extracted by the encoder ${\bm{\phi^{E}_{i}}}$ as $F_i\in\mathcal{R}^{C_i\times H_i\times W_i}$, where $C_i$, $H_i$, and $W_i$ represent channel, height, and weight, respectively. 
The encoding process can be represented as follows:
\begin{equation}
    \begin{aligned}
        F_i = {\bm{\phi^{E}_{i}}}(F_{i-1}),~~~i=1,2,\dots,N, 
    \end{aligned}
\end{equation}
where the resolution of {\bm{$\phi^{E}_{i}$}} equals {\bm{$\phi^{E}_{i-1}$}} if columnar encoder is employed, and vice versa.

\vspace{1mm}
\noindent{\textbf{\textit{2)} Feature Fuser.}} This module integrates multi-layer features $\{F_{0}, F_{1}, \dots, F_{N}\}$ from different stages and generates a more compact feature for the decoder. 
There are two main designs for this module: i) a light structure to fuse multi-scale features efficiently~\citep{uniad}; and ii) a heavy structure to obtain better representation~\citep{rd}. 
Without loss of generality, the Fuser {\textit{\bm{$\mathcal{F}$}}} takes multi-layer features as the input to obtain the fused feature $\hat{F}_{f}$, denoted as: 
\begin{equation}
    \begin{aligned}
        \hat{F}_{f} =  {\bm{\mathcal{F}}}(\theta_{0}(F_{0}), \theta_{1}(F_{1}), \dots, \theta_{N}(F_{N})), 
    \end{aligned}
    \label{eq:fuser_meta}
\end{equation}
where $\theta_{i}$ is used to adjust $i$-th feature to the desired size for subsequent fusion processing, \eg, up-sampling and de-convolution operations. 
Specifically, $\theta_{i}$ does not change the resolution of the $i$-th feature in the case of the columnar encoder. 
The model degenerates into an \textit{Auto Encoder} when only the last compressed feature $F_{N}$ is used, \ie, $\hat{F}_{f}$ equals $\hat{F}_{N}$.

\vspace{1mm}
\noindent{\textbf{\textit{3)} Feature Decoder.}} It reconstructs original images or encoded features from the fused feature $\hat{F}_{f}$. 
Similar to the encoder, the decoder {\bm{$\phi^{D}$}} can be a pyramidal/columnar structure. 
A pyramidal CNN is the most commonly used structure because the fused feature usually needs to be up-sampled to match the resolution of encoded features for accurate anomaly localization. 
The reconstruction process can be formulated as follows: 
\begin{equation}
    \begin{aligned}
        \hat{F}_{i-1} =  {\bm{\phi^{D}_{i}}}(\hat{F}_{i}),~~~i=1,2,\dots,N,
    \end{aligned}
\end{equation}
where the resolution of {\bm{$\phi^{D}_{i}$}} equals  {\bm{$\phi^{D}_{i-1}$}} if columnar encoder is employed, and vice versa.

\vspace{1mm}
\noindent{\textbf{\textit{4)} Anomaly Map Estimation.}} Reconstruction-based AD methods assume that anomalies cannot be well reconstructed by features extracted from normal images during the training process. 
For the $i$-th stage, the anomaly map $A_{i}$ is computed by:
\begin{equation}
    \begin{aligned}
        A_{i} = \mathcal{L}_{i} (F_{i}, \hat{F}_{i}),~~~i=1,2,\dots,N,
    \end{aligned}
\end{equation}
where $\mathcal{L}_{i}$ can be L1, Mean Square Error (MSE), Cosine Distance, and other metrics.
During the training phase, $A = \Sigma_{i}^{N}A_{i}$ is generally used as the loss to backward the gradient, while $A = \Sigma_{i}^{N}{A_{i}}$ serves as the anomaly map for testing. 
% $\lambda_{i}$ is the weight to balance different-depth anomaly maps, and we set it as 1 in the paper by default. 
Nevertheless, some reconstruction-based works employ extra losses~\citep{rrd,ocrgan} for further improvement, but they suffer from complicated modeling and implementation issues.
In this paper, our work uses only one simplest pixel-level loss sufficient for Meta-AD.

\vspace{1mm}
\noindent{\textbf{\textit{5)} Generalized Extension.}} In the generalized concept, encoder features can skip directly to multiple decoder stages without going through the Fuser~\citep{skipganomaly}. 
Nevertheless, this manner tends to fall into an "identity shortcut" that appears to return an unmodified input, disregarding its input content. However, it can be mitigated through specific training strategies and data augmentation methods~\citep{uniad}. 
Nevertheless, this paper does not discuss this manner, but the abstracted Meta-AD framework can be easily extended to include that case. 
In addition, effective plug-and-play defect construction~\citep{dtd,draem}, data augmentation~\citep{simplenet}, and loss functions~\citep{cutpaste,rrd} techniques can potentially enhance performance further.

\begin{figure}[tp]
    \centering
    \includegraphics[width=1.0\linewidth]{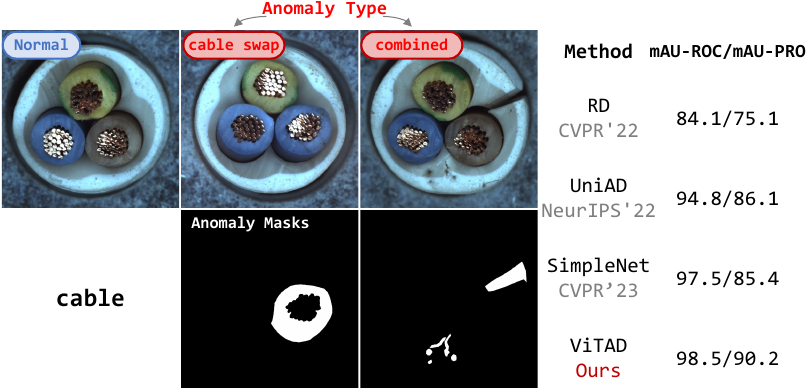}
    \caption{Left: Pilot study for the necessity of global dependence on "cable" category, \eg, logically-dependent "cable swap" and long-distance dependent "combined" defects. Right: Quantitative evaluations. Our ViTAD markedly mitigates these challenges.}
    \label{fig:cable}
\end{figure}

\begin{figure*}[tp]
    \centering
    \includegraphics[width=1.0\linewidth]{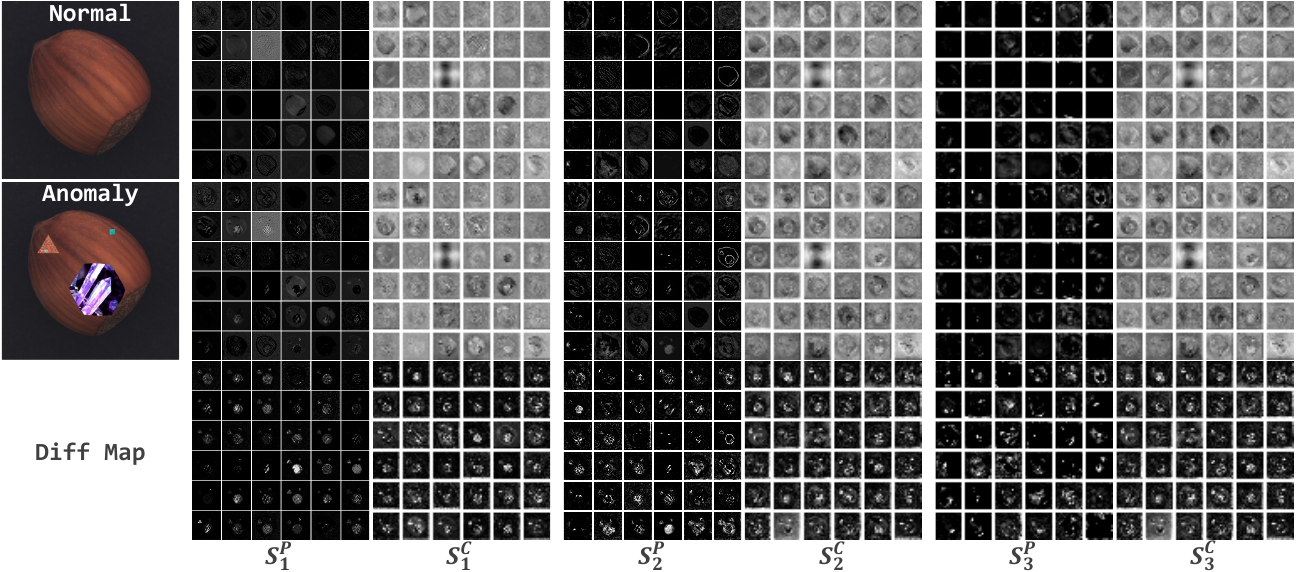}
    \caption{\textbf{First 36 visualized feature maps of different stages ($S_{i}$, $i=1, 2, 3$) for \textit{pyramid}~\citep{rd} ($S^{P}$) and \textit{columnar}~\citep{vit} ($S^{C}$) backbones.} The first two rows show the results of normal and anomaly images in the first column, and the last row shows differential maps. Results demonstrate the superiority of ViT for capturing more abundant features and locating more distinct anomalous regions.}
    \label{fig:feas_vis}
\end{figure*}

\subsection{Plain Vision Transformer for AD} \label{section:vitad}
\subsubsection{Motivation for exploring plain ViT for the MUAD task} \label{section:vitad_motivation}
Thanks to the global modeling capability and powerful feature representations, ViT exhibits remarkable achievements for diverse scenes, which has been proven effective and applied in various domains, \ie, object detection~\citep{vitdet}, semantic segmentation~\citep{sam}, in-context visual learning~\citep{seggpt}, and multi-modal fusion~\citep{vilt}. 
Nevertheless, \textit{using only plain ViT without additional structures has never been explored in the AD field.}
Recently, some AD methods~\citep{vt_adl,adtr,uniad} have introduced powerful ViT as part of the model (\eg, encoder or decoder) to improve performance. 
Although some progress has been made, \textit{relying on only a portion of global ViT modeling still leads to misdetection of certain types of defects like previous methods without ViT}, \eg, logical errors ("cable swap") and shortcomings in long-distance interactions ("combined") in cable category on MVTec AD~\citep{mvtec} in Fig.~\ref{fig:cable}.
This inspires us to explore the feasibility of using pure ViT for AD tasks. 
Specifically, we take a pair of normal and abnormal images for a toy experiment shown in \cref{fig:feas_vis}.
The ViT features are more prosperous and diversified than those from CNN at each stage, and the difference between normal and abnormal images is more significant. 
The phenomenon demonstrates the stronger modeling capability and larger receptive field of ViT structure for the potential applications to the AD task. 

% 
% 1) A ViT can extract features according to the input image content, which allows for stronger modeling capability and generalization compared to the static CNNs~\citep{deit,vitdet,sam}. 
%
% As shown in \cref{fig:feas_vis}, ViT features are richer and more diversified than CNN's at each stage, and the difference between normal and abnormal images is more significant. 
%
% 2) A ViT provides a larger receptive field, enabling it to better capture potential long-range correlations among distant positions. 
%
% Furthermore, ViT can model both high-/low-frequency information simultaneously, while CNN is more adept at high-frequency information~\citep{nextvit,iformer}.
%
% This enables ViTs to locate subtle defects, an important property for precise anomaly detection tasks. 
%

\subsubsection{Progressive Design of ViTAD} \label{section:vitad_global_local}
As multi-resolution features are necessary to model anomaly location accurately, existing AD methods use pyramidal networks in encoder~\citep{adtr,uniad} or decoder~\citep{vt_adl,destseg} to extract multi-resolution features. 
In contrast, we develop a non-pyramidal plain yet effective ViTAD for the MUAD task from Meta-AD step by step. 

\begin{figure}[tp]
    \centering
    \includegraphics[width=1.0\linewidth]{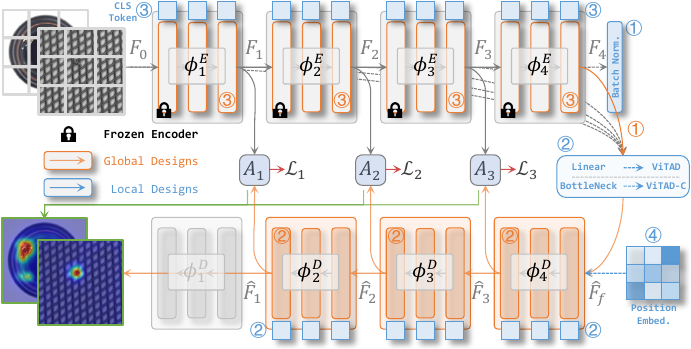}
    \caption{\textbf{Detailed structure of instantiated ViTAD from Meta-AD.} 
    Partial structures represented by the \textcolor{orange_vitad}{\textbf{orange}} and \textcolor{blue_vitad}{\textbf{blue}} numbers indicate the adaptations of the plain ViT at the global and local levels for anomaly detection.}
    \label{fig:vitad}
\end{figure}

Based on the formulation of Meta-AD, we employ columnar plain ViT as the instantiated structure and divide it into four stages as pyramid RD~\citep{rd} for both encoder $\{ {\bm{\phi^{E}_{1}}},  {\bm{\phi^{E}_{2}}},  {\bm{\phi^{E}_{3}}},  {\bm{\phi^{E}_{4}}}\}$ and decoder $\{ {\bm{\phi^{D}_{1}}},  {\bm{\phi^{D}_{2}}},  {\bm{\phi^{D}_{3}}},  {\bm{\phi^{D}_{4}}}\}$.
Each stage contains the same number of layers, \ie, 3 layers for each stage with 12-layer ViT-B trained on ImageNet-1K, and the decoder is randomly initialized with the same ViT structure. 
For the feature Fuser $ {\textit{\bm{\mathcal{F}}}}$, we employ one superficial linear layer to aggregate concatenated multi-stage features with the same resolution to ensure that the channel number of $\hat{F}_{f}$ matches the decoder input: 
\begin{equation}
    \begin{aligned}
        \hat{F}_{f} =  {\bm{\mathcal{F}}}(F_{1}, \dots, F_{4}) 
        = \text{Linear}([F_{1}, \dots, F_{4}]), 
    \end{aligned}
    \label{eq:fuser}
\end{equation}
where $\theta_{i}$ in \cref{eq:fuser_meta} degenerates into identity operation for simplicity. 
Nevertheless, a naive implementation yields a significant gap compared to the state-of-the-art methods for the MUAD task, \eg, the image-level and pixel-level mAU-ROC metrics are only 92.4/95.9, which are significantly lower than the SoTA UniAD of 97.5/97.0. 
Thus, we make AD-specific improvements to the implementation details at both global and local levels in \cref{fig:vitad}, ultimately achieving obvious superiority over SoTA pyramidal approaches (\cref{exp:sotas}). 

\noindent \textbf{Global Design of ViTAD.} 
As shown by the \textcolor{orange_vitad}{\textbf{orange}} numbers in \cref{fig:vitad}, we explore three effective designs at the global level for the ViT-based AD model. 
First, the Fuser removes the multi-scale feature skip connection and only uses the last stage $F_{4}$ as input, \ie, $\hat{F}_{f}=\text{Linear}(F_{4})$ for \cref{eq:fuser}. This modification significantly improves model performance (see \cref{tab:ab_fuser}), \eg, the mAD increases from 84.7 to 85.4. 
%
% \Eg, the mAU-ROC/mAP/mAP/m$F_1$-max metrics respectively increase by +0.7/+0.4/+0.9, while the pixel-level results remain almost unchanged (\cref{tab:ab_fuser}). 
%
The reason is that the deep features $F_{4}$ of the columnar ViT are sufficient to contain rich texture and semantic attributes. 
The injection of early features would shorten the information flow path, leading to potential information leakage that enables the model to learn the identity mapping~\citep{rd,uniad} and affecting the model's judgment ability at the image level. 
Furthermore, we observe that a heavy fuser is not necessary and a superficial linear layer suffices, which contradicts the design conclusion of previous works~\citep{HaloAE,rd} (\cref{exp:local}). 
Second, features at different levels can express fine-grained features differently. 
In this work, we use $F_{1}$/$F_{2}$/$F_{3}$ for training constraints and compute anomaly maps $A_{1}$/$A_{2}$/$A_{3}$ during inference. (see \cref{tab:ab_restraint})
This configuration differs from the default usage of the first three stages in works~\citep{rd,uniad}, providing richer multi-depth information to produce more accurate anomaly localization. 
Third, existing AD methods would use ImageNet-1K~\citep{imagenet} pretrained vision backbone as part of the model by default. 
However, this straightforward approach performs poorly due to the domain gap between ImageNet-1K and downstream AD datasets. 
We analyze various self-supervised training schemes, empirically observing that pretrained weights from more general unsupervised training could mitigate this gap. 
%
% As shown in \cref{exp:ab_pretrain}, larger supervised models and pre-training manners improve results (IN22K \vs IN1K). 
%
DINO~\citep{dino} has stronger semantic granularity and performs better than other pre-training methods (see \cref{tab:ab_pretrain}). 
%
% It is worth mentioning that MAE~\citep{mae} does not achieve the expected results, reasoning that the pre-training paradigm with a shallow decoder leads to weaker feature semantic expression ({\textbf{Finding~2}}). 
%
% Furthermore, we find that a smaller patch size (\ie, 8) and higher resolution significantly improve pixel-level indicators and overall mAD. 
%
% Although a slight decrease in image-level indicators accompanies this, improving segmentation results provides more valuable guidance for practical applications. 
%

%\subsubsection{Micro-level Modification of ViTAD} %\label{section:vitad_micro}
\noindent \textbf{Local Improvements of ViTAD.} To further improve performance, we analyze four structural details that potentially boost our model, as indicated by the \textcolor{blue_vitad}{\textbf{blue}} numbers in \cref{fig:vitad}. 
First, using features before normalization, rather than after this operation, as the fusion features slightly decrease image-level performance. 
Second, we use a lightweight linear one-layer as the Fuser structure that discards heavy transformations.
When we remove this linear layer, \ie, $F_4$=$\hat{F}_4$, it leads to performance loss, indicating the necessity of one linear layer.
% \red{({\textbf{Finding~3}}).} 
%
% Furthermore, we expand a more powerful ViTAD-C by empirically introducing a simple one-layer BottleNeck~\citep{resnet} in Fuser, further significantly improving all metrics. This suggests that the pure ViT-based structure can further benefit from the integration of a simple CNN-based module in Fuser. 
%
Third, maintaining the class token throughout the process slightly increases computational load and potentially affects the interaction between normal spatial structures, which causes a slight performance degradation. Therefore, we remove the class token, similar to previous works~\citep{swin,emo} in downstream task experiments. 
Fourth, maintaining the position embedding of the ViT-based decoder also brings slight performance gain.
Detailed ablation study can be viewed in \cref{tab:ab_details}. 

\subsubsection{ViTAD Summary} \label{section:summary}
Starting from the abstract Meta-AD framework, we gradually improve plain ViT from global and local perspectives to realize the adaptation to the AD task. 
As shown in \cref{tab:incremental}, each design would incrementally boost the performance, and we finally obtain a very competitive AD model for the challenging MUAD setting. 
We use the entry at the end of \cref{tab:incremental} as our final Meta-AD instantiation termed ViTAD, the first ViT-based powerful model customized for the AD domain.
The detailed structure can be viewed in the attached source code.

\subsubsection{Training Constraints} \label{section:vitad_constraints}
We aim to explore the effectiveness of plain ViT for the AD task and provide a strong baseline to facilitate future research. Therefore, we use only the cosine distance loss during training.

Specifically, ViTAD uses the fused feature $\hat{F}_f$ to reconstruct multiple features $F_{i}\in\mathcal{R}^{C_i\times H_i \times W_i}$ by the predicted $\hat{F}_{i}\in\mathcal{R}^{C_i\times H_i \times W_i}$ with decoder. 
Let $F_{i}(h, w)\in\mathcal{R}^{C_i}$ and $\hat{F}_{i}(h, w)\in\mathcal{R}^{C_i}$ be the $i$-th stage feature vector at position $(h, w)$ in both encoder and decoder, we use the cosine distance at position $(h, w)$ as the anomaly score $A_{i}(h, w)$~\citep{rd}: 
\begin{equation}
    \begin{aligned}
        A_i(h, w) = 1 - \frac{F_{i}(h, w)^T\cdot \hat{F}_{i}(h, w)}{\Vert F_{i}(h, w) \Vert \Vert \hat{F}_{i}(h, w) \Vert}, i \in \{1, 2, 3\}.
    \end{aligned}
\end{equation}
Anomaly scores of all positions construct the final anomaly map. 
Since the model is only trained on normal samples, the large pixel values in the anomaly map suggest anomalies.
For the $i$-th stage, the corresponding loss term $\mathcal{L}_{i}$ is calculated by:
\begin{equation}
    \begin{aligned}
        \mathcal{L}_{i}=\frac{1}{H_i W_i}\sum_{h=1}^{H_i}\sum_{w=1}^{W_i}A_i(h, w).
    \end{aligned}
\end{equation}
The overall loss $\mathcal{L}_{All} = \Sigma_{1}^{3}{\mathcal{L}_{i}}$ is the simple sum of all $\mathcal{L}_i$ terms.

\begin{table}[tp]
    \caption{\textbf{Overall incremental trajectory from naive ViT-adapted Meta-AD to ViTAD.} Each line is based on a modification of the immediately preceding line. Detailed ablations in \cref{exp:ablation}.}
    \renewcommand{\arraystretch}{1.2}
    \setlength\tabcolsep{1.0pt}
    \resizebox{1.0\linewidth}{!}{
        % \begin{tabular}{p{0.6cm}<{\centering} p{3.6cm}<{\raggedright} p{2.0cm}<{\centering} p{2.3cm}<{\centering} p{1.0cm}<{\centering} }
        \begin{tabular}{clccc}
        \toprule[1.5pt]
        \multicolumn{2}{c}{Method} & Image-level & Pixel-level & mAD  \\
        \hline
        & \pzo~Baseline & 92.4/96.4/94.2 & 95.9/46.9/51.8/86.5 & 80.6  \\
        \hline
        \multirow{3}{*}{\rotatebox{90}{Global}} & +~Remove multi-depth features & 93.3/96.8/94.8 & 96.1/47.6/52.6/86.9 & 81.2  \\
        & +~Use late three stage features & 93.8/97.3/95.2 & 96.4/48.6/53.6/87.6 & 81.8  \\
        & +~Use pretrained DINO weights & 97.6/99.0/96.8 & 97.5/55.0/58.2/91.0 & 85.0  \\
        \hline
        \multirow{4}{*}{\rotatebox{90}{Local}} & \pzo~+~Before normalization & 97.9/99.1/97.0 & 97.5/54.8/58.1/91.1 & 85.1  \\
        & \pzo~+~Add linear & 98.1/99.2/97.0 & 97.6/55.1/58.6/91.4 & 85.3  \\
        & \pzo~+~Remove class token & 98.0/99.1/97.2 & 97.7/55.3/58.5/91.3 & 85.3  \\
        & \pzo~+~Use position embedding & 98.3/99.4/97.3 & 97.7/55.3/58.7/91.4 & 85.4  \\
        \bottomrule[1.5pt]
        \end{tabular}
    }
    \label{tab:incremental}
\end{table}

\subsubsection{Anomaly Segmentation and Classification} \label{section:vitad_inference}
Anomaly segmentation aims to provide the pixel-level anomaly score map to determine the specific locations of the anomalies. 
Multi-scale anomaly maps $\{A_1, A_2, A_3\}$ calculated by the restrained features are summed up to form the final anomaly map $A$=$\text{sum}(A_1, A_2, A_3)$ for evaluation. 
In addition, anomaly classification requires an image-level anomaly score to indicate whether the image is anomalous. 
Similar to~\citep{uniad}, we first apply an average pooling operation to the anomaly score map $A$ and then take its maximum value as the anomaly score.

\begin{table*}[htp]
    \centering
    \caption{Comparison for concurrent powerful AD methods. \cmark: Satisfied; \xmark: Unsatisfied; \pmark: Partially satisfied; \omark: Inapplicable.}
    \label{tab:model_criterion}
    \renewcommand{\arraystretch}{1.2}
    \setlength\tabcolsep{3.0pt}
    \resizebox{1.0\linewidth}{!}{
        % \begin{tabular}{p{1.7cm}<{\centering} p{1.2cm}<{\centering} p{1.2cm}<{\centering} p{1.2cm}<{\centering} p{2.0cm}<{\centering} p{2.0cm}<{\centering} p{0.7cm}<{\centering} p{0.7cm}<{\centering} p{0.7cm}<{\centering} p{4.0cm}<{\raggedright}}
        \begin{tabular}{cccccccccl}
            %\toprule[0.17em]
            \toprule 
            Criterion~$\rightarrow$ & \multirow{2}{*}{\makecell[c]{Pyramidal\\ Encoder}} & \multirow{2}{*}{\makecell[c]{Heavy\\ Fuser}} & \multirow{2}{*}{\makecell[c]{Pyramidal\\ Decoder}} & \multirow{2}{*}{\makecell[c]{Multi-Resolution\\ Features}} & \multirow{2}{*}{\makecell[c]{Image/Feature\\ Augmentation}} & \multicolumn{3}{c}{Category} & \multirow{2}{*}{\makecell[c]{\pzo\pzo\pzo\pzo\pzo\pzo\pzo\pzo Reproduction\\ \pzo\pzo\pzo\pzo\pzo\pzo\pzo\pzo Code}} \\
            \cline{1-1} \cline{7-9}
            Method~$\downarrow$ & & & & & & Aug. & Emb. & Rec. & \\ 
            \hline
            DRAEM~\citep{draem} & \cmark & \xmark & \omark & \cmark & \cmark & \cmark & \pmark & \pmark & ~\href{https://github.com/VitjanZ/DRAEM}{github.com/VitjanZ/DRAEM} \\
            RD~\citep{rd}  & \cmark & \cmark & \cmark & \cmark & \xmark & \xmark & \xmark & \cmark & ~\href{https://github.com/hq-deng/RD4AD}{github.com/hq-deng/RD4AD} \\
            UniAD~\citep{uniad}  & \cmark & \xmark & \xmark & \cmark & \cmark & \cmark & \xmark & \cmark & ~\href{https://github.com/zhiyuanyou/UniAD}{github.com/zhiyuanyou/UniAD} \\
            DeSTSeg~\citep{destseg}  & \cmark & \xmark & \omark & \cmark & \cmark & \cmark & \cmark & \xmark & ~\href{https://github.com/apple/ml-destseg}{github.com/apple/ml-destseg} \\
            SimpleNet~\citep{simplenet}  & \cmark & \xmark & \omark & \cmark & \cmark & \cmark & \cmark & \xmark & ~\href{https://github.com/DonaldRR/SimpleNet}{github.com/DonaldRR/SimpleNet} \\
            \hline
            \rowcolor{tab_ours} ViTAD (Ours) & \xmark & \xmark & \xmark & \xmark & \xmark & \xmark & \xmark & \cmark & ~\href{https://github.com/zhangzjn/ader}{github.com/zhangzjn/ADer} \\
            %\toprule[0.12em]
            \bottomrule
        \end{tabular}
    }
\end{table*}

\section{Experiments} \label{section:exp}

\subsection{Setups for Multi-class Unsupervised AD} \label{section:setup}
\noindent\textbf{Task Setting.} 
The single-class setting usually requires training a model for each class separately.
This work focuses on the more challenging and practical multi-class setting described in \cref{section:muad}. 
Experiments for ablation and interpretability are primarily conducted on the MVTec AD dataset.

\vspace{1mm}
\noindent\textbf{Datasets.} 
We evaluate ViTAD and state-of-the-art methods on the widely-used industrial MVTec AD~\citep{mvtec,mvtec_ijcv} and VisA~\citep{visa} datasets for anomaly classification and segmentation: 
The MVTec AD dataset contains 15 industrial products in 2 types, with 3,629 normal images for training and 467/1,258 normal/anomaly images for testing (5,354 images in total); 
The VisA dataset covers 12 objects in 3 types, with 8,659 normal images for training and 962/1,200 normal/anomaly images for testing (10,821 images in total). 
In addition, we unify Brain MRI (7,500 normal images for training and 640/3,075 normal/anomaly images for testing), Liver CT (1,542 normal images for training and 660/1,258 normal/anomaly images for testing), and Retinal OCT (4,297 normal images for training and 1,041/764 normal/anomaly images for testing) to establish a Uni-Medical~\citep{bmad} benchmark in the medical field.
All datasets provide ground-truth anomaly maps at the pixel level for evaluation.

\begin{table*}[thb]
    \centering
    \caption{\textbf{Image-level multi-class anomaly classification results with mAU-ROC/mAP/m$F_1$-max metrics on MVTec AD.} Note that only one model is trained to detect anomalies for all categories. \textbf{Superscript $^*$: Re-trained under the multi-class setting by the official code. $^\dagger$: Reproduced results no less than the original paper}. \textbf{\textit{Bold}} and \underline{\textit{Underline}} indicate optimal and sub-optimal results, respectively. The circled number represents the category to which the method belongs (\cf, \cref{section:related}): \textbf{\ding{192}}-Image/Feature augmentation based; \textbf{\ding{193}}-Embedding based; \textbf{\ding{194}}-Image/Feature Reconstruction based, and the \red{\textbf{red}} circle indicate the main category to which the method belongs. \textbf{\textit{Subsequent tables follow the consistent presentations.}}}
    \label{tab:mvtec_sp}
    \renewcommand{\arraystretch}{1.1}
    \setlength\tabcolsep{3.0pt}
    \resizebox{1.0\linewidth}{!}{
        % \begin{tabular}{p{0.2cm}<{\centering} p{2.0cm}<{\centering} p{2.0cm}<{\centering} p{2.0cm}<{\centering} p{2.0cm}<{\centering} p{2.0cm}<{\centering} p{2.0cm}<{\centering} p{3.6cm}<{\centering}}
        \begin{tabular}{cccccccc}
            \toprule
            % \multicolumn{2}{c}{\multirow{2}{*}{Method~$\rightarrow$}} & \red{\textbf{\ding{192}}}~\textbf{\ding{193}}~\textbf{\ding{194}} & \red{\textbf{\ding{194}}} & \textbf{\ding{192}}~\red{\textbf{\ding{194}}} & \textbf{\ding{192}}~\red{\textbf{\ding{193}}} & \textbf{\ding{192}}~\red{\textbf{\ding{193}}} & \cellcolor{tab_ours}{\red{\textbf{\ding{194}}}} \\
            % & & DRAEM$^*$~\citep{draem} & RD$^*$~\citep{rd} & UniAD$^\dagger$~\citep{uniad} & DeSTSeg$^*$~\citep{destseg} & SimpleNet$^*$~\citep{simplenet} & \cellcolor{tab_ours}{ViTAD} \\ 
            % \cline{1-2}
            % \multicolumn{2}{c}{Category~$\downarrow$} & ICCV'21 & CVPR'22 & NeurIPS'22 & CVPR'23 & CVPR'23 & \cellcolor{tab_ours}{(Ours)} \\
            \multicolumn{2}{c}{\multirow{2}{*}{Method~$\rightarrow$}} & \red{\textbf{\ding{192}}}~\textbf{\ding{193}}~\textbf{\ding{194}} & \red{\textbf{\ding{194}}} & \textbf{\ding{192}}~\red{\textbf{\ding{194}}} & \textbf{\ding{192}}~\red{\textbf{\ding{193}}} & \textbf{\ding{192}}~\red{\textbf{\ding{193}}} & \cellcolor{tab_ours}{\red{\textbf{\ding{194}}}} \\
            & & \makecell[c]{DRAEM$^*$\\~\citep{draem}} & \makecell[c]{RD$^*$\\~\citep{rd}} & \makecell[c]{UniAD$^\dagger$\\~\citep{uniad}} & \makecell[c]{DeSTSeg$^*$\\~\citep{destseg}} & \makecell[c]{SimpleNet$^*$\\~\citep{simplenet}} & \cellcolor{tab_ours}{ViTAD} \\ 
            \cline{1-2}
            \multicolumn{2}{c}{Category~$\downarrow$} & ICCV'21 & CVPR'22 & NeurIPS'22 & CVPR'23 & CVPR'23 & \cellcolor{tab_ours}{(Ours)} \\
            \hline
            \multirow{5}{*}{\rotatebox{90}{Texture}} & Carpet & 97.2/99.1/96.7 & 98.5/\underline{99.6}/\underline{97.2} & \textbf{99.8}/\textbf{99.9}/\textbf{99.4} & 97.6/99.3/96.6 & 95.9/98.8/94.9 & \cellcolor{tab_ours}{\underline{99.5}{\tiny{\tr{$\pm.00$}}}/\textbf{99.9}{\tiny{\tr{$\pm.00$}}}/\textbf{99.4}{\tiny{\tr{$\pm.00$}}}} \\
            & \cellcolor{tab_others}{Grid} & \cellcolor{tab_others}{99.2/99.7/\underline{98.2}} & \cellcolor{tab_others}{98.0/99.4/96.6} & \cellcolor{tab_others}{\underline{99.3}/\underline{99.8}/\textbf{99.1}} & \cellcolor{tab_others}{97.9/99.2/96.6} & \cellcolor{tab_others}{97.6/99.2/96.4} & \cellcolor{tab_ours}{\textbf{99.7}{\tiny{\tr{$\pm.14$}}}/\textbf{99.9}{\tiny{\tr{$\pm.05$}}}/\textbf{99.1}{\tiny{\tr{$\pm.50$}}}} \\
            & Leather & 97.7/99.3/95.0 & \textbf{100.}/\textbf{100.}/\textbf{100.} & \textbf{100.}/\textbf{100.}/\textbf{100.} & \underline{99.2}/\underline{99.8}/\underline{98.9} & \textbf{100.}/\textbf{100.}/\textbf{100.} & \cellcolor{tab_ours}{\textbf{100.}{\tiny{\tr{$\pm.00$}}}/\textbf{100.}{\tiny{\tr{$\pm.00$}}}/\textbf{100.}{\tiny{\tr{$\pm.00$}}}} \\
            & \cellcolor{tab_others}{Tile} & \cellcolor{tab_others}{\textbf{100.}/\textbf{100.}/\textbf{100.}} & \cellcolor{tab_others}{98.3/99.3/96.4} & \cellcolor{tab_others}{\underline{99.9}/\underline{99.9}/\underline{99.4}} & \cellcolor{tab_others}{97.0/98.9/95.3} & \cellcolor{tab_others}{99.3/99.8/98.8} & \cellcolor{tab_ours}{\textbf{100.}{\tiny{\tr{$\pm.00$}}}/\textbf{100.}{\tiny{\tr{$\pm.00$}}}/\textbf{100.}{\tiny{\tr{$\pm.00$}}}} \\
            & Wood & \textbf{100.}/\textbf{100.}/\textbf{100.} & 99.2/\underline{99.8}/98.3 & 98.9/99.7/97.5 & \underline{99.9}/\textbf{100.}/\underline{99.2} & 98.4/99.5/96.7 & \cellcolor{tab_ours}{98.7{\tiny{\tr{$\pm.10$}}}/99.6{\tiny{\tr{$\pm.03$}}}/96.7{\tiny{\tr{$\pm.47$}}}} \\
            \hline
            \multirow{10}{*}{\rotatebox{90}{Object}}& Bottle & 97.3/99.2/96.1 & \underline{99.6}/\underline{99.9}/\underline{98.4} & \textbf{100.}/\textbf{100.}/\textbf{100.} & 98.7/99.6/96.8 & \textbf{100.}/\textbf{100.}/\textbf{100.} & \cellcolor{tab_ours}{\textbf{100.}{\tiny{\tr{$\pm.00$}}}/\textbf{100.}{\tiny{\tr{$\pm.00$}}}/\textbf{100.}{\tiny{\tr{$\pm.00$}}}} \\
            & \cellcolor{tab_others}{Cable} & \cellcolor{tab_others}{61.1/74.0/76.3} & \cellcolor{tab_others}{84.1/89.5/82.5} & \cellcolor{tab_others}{94.8/97.0/90.7} & \cellcolor{tab_others}{89.5/94.6/85.9} & \cellcolor{tab_others}{\underline{97.5}/\underline{98.5}/\underline{94.7}} & \cellcolor{tab_ours}{\textbf{98.5}{\tiny{\tr{$\pm.15$}}}/\textbf{99.1}{\tiny{\tr{$\pm.08$}}}/\textbf{95.7}{\tiny{\tr{$\pm.86$}}}} \\
            & Capsule & 70.9/92.5/90.5 & \underline{94.1}/96.9/\textbf{96.9} & 93.7/\underline{98.4}/\underline{96.3} & 82.8/95.9/92.6 & 90.7/97.9/93.5 & \cellcolor{tab_ours}{\textbf{95.4}{\tiny{\tr{$\pm.46$}}}/\textbf{99.0}{\tiny{\tr{$\pm.12$}}}/95.5{\tiny{\tr{$\pm.05$}}}} \\
            & \cellcolor{tab_others}{Hazelnut} & \cellcolor{tab_others}{94.7/97.5/92.3} & \cellcolor{tab_others}{60.8/69.8/86.4} & \cellcolor{tab_others}{\textbf{100.}/\textbf{100.}/\textbf{100.}} & \cellcolor{tab_others}{98.8/99.2/98.6} & \cellcolor{tab_others}{\underline{99.9}/\textbf{100.}/\underline{99.3}} & \cellcolor{tab_ours}{99.8{\tiny{\tr{$\pm.13$}}}/\underline{99.9}{\tiny{\tr{$\pm.07$}}}/98.6{\tiny{\tr{$\pm.82$}}}} \\
            & Metal Nut & 81.8/95.0/92.0 & \textbf{100.}/\textbf{100.}/\textbf{99.5} & 98.3/99.5/\underline{98.4} & 92.9/98.4/92.2 & 96.9/99.3/96.1 & \cellcolor{tab_ours}{\underline{99.7}{\tiny{\tr{$\pm.07$}}}/\underline{99.9}{\tiny{\tr{$\pm.02$}}}/\underline{98.4}{\tiny{\tr{$\pm.30$}}}} \\
            & \cellcolor{tab_others}{Pill} & \cellcolor{tab_others}{76.2/94.9/92.5} & \cellcolor{tab_others}{\textbf{97.5}/\textbf{99.6}/\textbf{96.8}} & \cellcolor{tab_others}{94.4/99.0/95.4} & \cellcolor{tab_others}{77.1/94.4/91.7} & \cellcolor{tab_others}{88.2/97.7/92.5} & \cellcolor{tab_ours}{\underline{96.2}{\tiny{\tr{$\pm.50$}}}/\underline{99.3}{\tiny{\tr{$\pm.10$}}}/96.4{\tiny{\tr{$\pm.18$}}}} \\
            & Screw & 87.7/95.7/89.9 & \textbf{97.7}/\textbf{99.3}/\textbf{95.8} & \underline{95.3}/\underline{98.5}/92.9 & 69.9/88.4/85.4 & 76.7/90.5/87.7 & \cellcolor{tab_ours}{91.3{\tiny{\tr{$\pm.39$}}}/97.0{\tiny{\tr{$\pm.16$}}}/\underline{93.0}{\tiny{\tr{$\pm.88$}}}} \\
            & \cellcolor{tab_others}{Toothbrush} & \cellcolor{tab_others}{90.8/96.8/90.0} & \cellcolor{tab_others}{\underline{97.2}/\underline{99.0}/94.7} & \cellcolor{tab_others}{89.7/95.3/\underline{95.2}} & \cellcolor{tab_others}{71.7/89.3/84.5} & \cellcolor{tab_others}{89.7/95.7/92.3} & \cellcolor{tab_ours}{\underline{98.9}{\tiny{\tr{$\pm.32$}}}/\underline{99.6}{\tiny{\tr{$\pm.12$}}}/\underline{96.8}{\tiny{\tr{$\pm.92$}}}} \\
            & Transistor & 77.2/77.4/71.1 & 94.2/95.2/90.0 & \textbf{99.8}/\textbf{99.8}/\underline{97.5} & 78.2/79.5/68.8 & \underline{99.2}/\underline{98.7}/\textbf{97.6} & \cellcolor{tab_ours}{98.8{\tiny{\tr{$\pm.57$}}}/98.3{\tiny{\tr{$\pm.85$}}}/92.5{\tiny{\tr{$\pm.72$}}}} \\
            & \cellcolor{tab_others}{Zipper} & \cellcolor{tab_others}{\textbf{99.9}/\textbf{100.}/\textbf{99.2}} & \cellcolor{tab_others}{\underline{99.5}/\underline{99.9}/\textbf{99.2}} & \cellcolor{tab_others}{98.6/99.6/97.1} & \cellcolor{tab_others}{88.4/96.3/93.1} & \cellcolor{tab_others}{99.0/99.7/\underline{98.3}} & \cellcolor{tab_ours}{97.6{\tiny{\tr{$\pm.08$}}}/99.3{\tiny{\tr{$\pm.03$}}}/97.1{\tiny{\tr{$\pm.46$}}}} \\
            \hline
            \multicolumn{2}{c}{Average} & 88.8/94.7/92.0 & 94.6/96.5/95.2 & \underline{97.5}/\underline{99.1}/\textbf{97.3} & 89.2/95.5/91.6 & 95.3/98.4/\underline{95.8} & \cellcolor{tab_ours}{\textbf{98.3}{\tiny{\tr{$\pm.02$}}}/\textbf{99.4}{\tiny{\tr{$\pm.05$}}}/\textbf{97.3}{\tiny{\tr{$\pm.20$}}}} \\
            \bottomrule
        \end{tabular}
    }
    \vspace{-1.0em}
\end{table*}

\begin{table*}[thb]
    \centering
    \caption{\textbf{Pixel-level multi-class anomaly segmentation with mAU-ROC/mAP/m$F_1$-max/mAU-PRO on MVTec AD.} The last row presents the averaged mAD metric across seven metrics to provide a comprehensive evaluation.
    }
    \label{tab:mvtec_px}
    \renewcommand{\arraystretch}{1.2}
    \setlength\tabcolsep{1.0pt}
    \resizebox{1.0\linewidth}{!}{
        % \begin{tabular}{p{0.2cm}<{\centering} p{2.0cm}<{\centering} p{2.7cm}<{\centering} p{2.7cm}<{\centering} p{2.7cm}<{\centering} p{2.7cm}<{\centering} p{2.7cm}<{\centering} p{4.7cm}<{\centering}}
        \begin{tabular}{cccccccc}
            \toprule
            % \multicolumn{2}{c}{Method~$\rightarrow$} & DRAEM$^*$~\citep{draem} & RD$^*$~\citep{rd} & UniAD$^\dagger$~\citep{uniad} & DeSTSeg$^*$~\citep{destseg} & SimpleNet$^*$~\citep{simplenet} & \cellcolor{tab_ours}{ViTAD} \\ 
            \multicolumn{2}{c}{Method~$\rightarrow$} & \makecell[c]{DRAEM$^*$\\~\citep{draem}} & \makecell[c]{RD$^*$\\~\citep{rd}} & \makecell[c]{UniAD$^\dagger$\\~\citep{uniad}} & \makecell[c]{DeSTSeg$^*$\\~\citep{destseg}} & \makecell[c]{SimpleNet$^*$\\~\citep{simplenet}} & \cellcolor{tab_ours}{ViTAD} \\ 
            \cline{1-2}
            \multicolumn{2}{c}{Category~$\downarrow$} & ICCV'21 & CVPR'22 & NeurIPS'22 & CVPR'23 & CVPR'23 & \cellcolor{tab_ours}{(Ours)} \\
            \hline
            \multirow{5}{*}{\rotatebox{90}{Texture}} & Carpet & 98.1/\textbf{78.7}/\textbf{73.1}/93.1 & \textbf{99.0}/58.5/60.5/\textbf{95.1} & \underline{98.4}/51.4/51.5/94.4 & 93.6/59.9/58.9/89.3 & 97.4/38.7/43.2/90.6 & \cellcolor{tab_ours}{\textbf{99.0}{\tiny{\tr{$\pm.01$}}}/\underline{60.5}{\tiny{\tr{$\pm.60$}}}/\underline{64.1}{\tiny{\tr{$\pm.22$}}}/\underline{94.7}{\tiny{\tr{$\pm.19$}}}} \\
            & \cellcolor{tab_others}{Grid} & \cellcolor{tab_others}{\underline{99.0}/\underline{44.5}/46.2/92.1} & \cellcolor{tab_others}{\textbf{99.2}/\textbf{46.0}/\textbf{47.4}/\textbf{97.0}} & \cellcolor{tab_others}{97.7/23.7/30.4/92.9} & \cellcolor{tab_others}{97.0/42.1/\underline{46.9}/86.8} & \cellcolor{tab_others}{96.8/20.5/27.6/88.6} & \cellcolor{tab_ours}{98.6{\tiny{\tr{$\pm.01$}}}/31.2{\tiny{\tr{$\pm.16$}}}/36.7{\tiny{\tr{$\pm.12$}}}/\underline{95.8}{\tiny{\tr{$\pm.82$}}}} \\
            & Leather & 98.9/\underline{60.3}/\underline{57.4}/88.5 & 99.3/38.0/45.1/\underline{97.4} & 98.8/34.2/35.5/96.8 & \underline{99.5}/\textbf{71.5}/\textbf{66.5}/91.1 & 98.7/28.5/32.9/92.7 & \cellcolor{tab_ours}{\textbf{99.6}{\tiny{\tr{$\pm.01$}}}/52.1{\tiny{\tr{$\pm.76$}}}/55.8{\tiny{\tr{$\pm.66$}}}/\textbf{97.9}{\tiny{\tr{$\pm.30$}}}} \\
            & \cellcolor{tab_others}{Tile} & \cellcolor{tab_others}{\textbf{99.2}/\textbf{93.6}/\textbf{86.0}/\textbf{97.0}} & \cellcolor{tab_others}{95.3/48.5/60.5/85.8} & \cellcolor{tab_others}{92.3/41.5/50.3/78.4} & \cellcolor{tab_others}{93.0/\underline{71.0}/66.2/87.1} & \cellcolor{tab_others}{95.7/60.5/59.9/\underline{90.6}} & \cellcolor{tab_ours}{\underline{96.6}{\tiny{\tr{$\pm.01$}}}/56.4{\tiny{\tr{$\pm.26$}}}/\underline{68.8}{\tiny{\tr{$\pm.14$}}}/87.0{\tiny{\tr{$\pm.44$}}}} \\
            & Wood & \textbf{96.9}/\textbf{81.4}/\textbf{74.6}/\textbf{94.2} & 95.3/47.8/51.0/\underline{90.0} & 93.2/37.4/42.8/86.7 & 95.9/\underline{77.3}/\underline{71.3}/83.4 & 91.4/34.8/39.7/76.3 & \cellcolor{tab_ours}{\underline{96.4}{\tiny{\tr{$\pm.03$}}}/60.6{\tiny{\tr{$\pm.52$}}}/58.3{\tiny{\tr{$\pm.39$}}}/88.0{\tiny{\tr{$\pm.69$}}}} \\
            \hline
            \multirow{10}{*}{\rotatebox{90}{Object}}& Bottle & 91.3/62.5/56.9/80.7 & 97.8/\underline{68.2}/67.6/\underline{94.0} & \underline{98.0}/67.0/\underline{67.9}/93.1 & 93.3/61.7/56.0/67.5 & 97.2/53.8/62.4/89.0 & \cellcolor{tab_ours}{\textbf{98.8}{\tiny{\tr{$\pm.03$}}}/\textbf{79.9}{\tiny{\tr{$\pm.57$}}}/\textbf{75.6}{\tiny{\tr{$\pm.56$}}}/\textbf{94.3}{\tiny{\tr{$\pm.12$}}}} \\
            & \cellcolor{tab_others}{Cable} & \cellcolor{tab_others}{75.9/14.7/17.8/40.1} & \cellcolor{tab_others}{85.1/26.3/33.6/75.1} & \cellcolor{tab_others}{\textbf{96.9}/\textbf{45.4}/\underline{50.4}/\underline{86.1}} & \cellcolor{tab_others}{89.3/37.5/40.5/49.4} & \cellcolor{tab_others}{\underline{96.7}/42.4/\textbf{51.2}/85.4} & \cellcolor{tab_ours}{96.2{\tiny{\tr{$\pm.25$}}}/\underline{43.1}{\tiny{\tr{$\pm.54$}}}/47.4{\tiny{\tr{$\pm.65$}}}/\textbf{90.2}{\tiny{\tr{$\pm.74$}}}} \\
            & Capsule & 50.5/\pzo6.0/10.0/27.3 & \textbf{98.8}/43.4/\textbf{50.1}/\textbf{94.8} & \textbf{98.8}/\underline{45.6}/47.7/\underline{92.1} & 95.8/\textbf{47.9}/\underline{48.9}/62.1 & \underline{98.5}/35.4/44.3/84.5 & \cellcolor{tab_ours}{98.3{\tiny{\tr{$\pm.02$}}}/42.7{\tiny{\tr{$\pm.58$}}}/47.8{\tiny{\tr{$\pm.28$}}}/92.0{\tiny{\tr{$\pm.94$}}}} \\
            & \cellcolor{tab_others}{Hazelnut} & \cellcolor{tab_others}{96.5/\textbf{70.0}/60.5/78.7} & \cellcolor{tab_others}{97.9/36.2/51.6/92.7} & \cellcolor{tab_others}{98.0/53.8/56.3/\underline{94.1}} & \cellcolor{tab_others}{98.2/\underline{65.8}/\underline{61.6}/84.5} & \cellcolor{tab_others}{\underline{98.4}/44.6/51.4/87.4} & \cellcolor{tab_ours}{\textbf{99.0}{\tiny{\tr{$\pm.01$}}}/64.6{\tiny{\tr{$\pm.46$}}}/\textbf{64.0}{\tiny{\tr{$\pm.25$}}}/\textbf{95.2}{\tiny{\tr{$\pm.08$}}}} \\
            & Metal Nut & 74.4/31.1/21.0/66.4 & 93.8/62.3/65.4/\underline{91.9} & 93.3/50.9/63.6/81.8 & 84.2/42.0/22.8/53.0 & \textbf{98.0}/\textbf{83.1}/\textbf{79.4}/85.2 & \cellcolor{tab_ours}{\underline{96.4}{\tiny{\tr{$\pm.14$}}}/\underline{75.1}{\tiny{\tr{$\pm.52$}}}/\underline{77.3}{\tiny{\tr{$\pm.83$}}}/\textbf{92.4}{\tiny{\tr{$\pm.13$}}}} \\
            & \cellcolor{tab_others}{Pill} & \cellcolor{tab_others}{93.9/59.2/44.1/53.9} & \cellcolor{tab_others}{\underline{97.5}/63.4/65.2/\textbf{95.8}} & \cellcolor{tab_others}{96.1/44.5/52.4/\underline{95.3}} & \cellcolor{tab_others}{96.2/61.7/41.8/27.9} & \cellcolor{tab_others}{96.5/\underline{72.4}/\underline{67.7}/81.9} & \cellcolor{tab_ours}{\textbf{98.7}{\tiny{\tr{$\pm.01$}}}/\textbf{77.8}{\tiny{\tr{$\pm.27$}}}/\textbf{75.2}{\tiny{\tr{$\pm.23$}}}/\underline{95.3}{\tiny{\tr{$\pm.15$}}}} \\
            & Screw & 90.0/33.8/40.7/55.2 & \textbf{99.4}/\textbf{40.2}/\textbf{44.7}/\textbf{96.8} & \underline{99.2}/\underline{37.4}/\underline{42.3}/\underline{95.2} & 93.8/19.9/25.3/47.3 & 96.5/15.9/23.2/84.0 & \cellcolor{tab_ours}{99.0{\tiny{\tr{$\pm.03$}}}/34.0{\tiny{\tr{$\pm.81$}}}/41.0{\tiny{\tr{$\pm.95$}}}/93.5{\tiny{\tr{$\pm.53$}}}} \\
            & \cellcolor{tab_others}{Toothbrush} & \cellcolor{tab_others}{97.3/\textbf{55.2}/55.8/68.9} & \cellcolor{tab_others}{\underline{99.0}/\underline{53.6}/\underline{58.8}/\textbf{92.0}} & \cellcolor{tab_others}{98.4/37.8/49.1/87.9} & \cellcolor{tab_others}{96.2/52.9/\underline{58.8}/30.9} & \cellcolor{tab_others}{98.4/46.9/52.5/87.4} & \cellcolor{tab_ours}{\textbf{99.1}{\tiny{\tr{$\pm.04$}}}/51.3{\tiny{\tr{$\pm.95$}}}/\textbf{61.9}{\tiny{\tr{$\pm.50$}}}/\underline{90.9}{\tiny{\tr{$\pm.17$}}}} \\
            & Transistor & 68.0/23.6/15.1/39.0 & 85.9/42.3/45.2/74.7 & \textbf{97.4}/\textbf{61.2}/\textbf{63.0}/\textbf{93.5} & 73.6/38.4/39.2/43.9 & \underline{95.8}/58.2/\underline{56.0}/\underline{83.2} & \cellcolor{tab_ours}{93.9{\tiny{\tr{$\pm.13$}}}/\underline{58.4}{\tiny{\tr{$\pm.36$}}}/55.3{\tiny{\tr{$\pm.42$}}}/76.8{\tiny{\tr{$\pm.88$}}}} \\
            & \cellcolor{tab_others}{Zipper} & \cellcolor{tab_others}{\textbf{98.6}/\textbf{74.3}/\textbf{69.3}/91.9} & \cellcolor{tab_others}{\underline{98.5}/53.9/\underline{60.3}/\textbf{94.1}} & \cellcolor{tab_others}{98.0/45.0/51.9/\underline{92.6}} & \cellcolor{tab_others}{97.3/\underline{64.7}/59.2/66.9} & \cellcolor{tab_others}{97.9/53.4/54.6/90.7} & \cellcolor{tab_ours}{95.9{\tiny{\tr{$\pm.07$}}}/42.6{\tiny{\tr{$\pm.43$}}}/50.8{\tiny{\tr{$\pm.30$}}}/87.2{\tiny{\tr{$\pm.82$}}}} \\
            \hline
            \multicolumn{2}{c}{Average} & 88.6/52.6/48.6/71.1 & 96.1/48.6/\underline{53.8}/\underline{91.2} & \textbf{97.0}/45.1/50.4/90.7 & 93.1/\underline{54.3}/50.9/64.8 & 96.9/45.9/49.7/86.5 & \cellcolor{tab_ours}{\textbf{97.7}{\tiny{\tr{$\pm.02$}}}/55.3{\tiny{\tr{$\pm.11$}}}/58.7{\tiny{\tr{$\pm.10$}}}/\underline{91.4}{\tiny{\tr{$\pm.21$}}}} \\
            \hline
            \multicolumn{2}{c}{mAD} & 76.6 & 82.3 & \underline{82.4} & 77.1 & 81.2 & \cellcolor{tab_ours}{\textbf{85.4}} \\
            \bottomrule
        \end{tabular}
    }
\end{table*}

\begin{table*}[thb]
    \centering
    \caption{\textbf{Multi-class image-level anomaly classification and pixel-level anomaly segmentation results on VisA.}}
    \label{tab:visa}
    \renewcommand{\arraystretch}{1.2}
    \setlength\tabcolsep{1.0pt}
    \resizebox{1.0\linewidth}{!}{
        % \begin{tabular}{p{0.56cm}<{\centering} p{2.3cm}<{\centering} p{2.0cm}<{\centering} p{2.0cm}<{\centering} p{2.0cm}<{\centering} p{3.6cm}<{\centering} p{0.1cm}<{\centering} p{2.7cm}<{\centering} p{2.7cm}<{\centering} p{2.7cm}<{\centering} p{4.7cm}<{\centering}}
        \begin{tabular}{ccccccccccc}
            \toprule
            & & \multicolumn{4}{c}{\textbf{Image-level} mAU-ROC/mAP/m$F_1$-max} & & \multicolumn{4}{c}{\textbf{Pixel-level} mAU-ROC/mAP/m$F_1$-max/mAU-PRO} \\
            \cline{3-6} \cline{8-11}
            % \multicolumn{2}{c}{Method~$\rightarrow$} & DRAEM$^*$~\citep{draem} & UniAD$^\dagger$~\citep{uniad} & SimpleNet$^*$~\citep{simplenet} & \cellcolor{tab_ours}{ViTAD} & & DRAEM$^*$~\citep{draem} & UniAD$^\dagger$~\citep{uniad} & SimpleNet$^*$~\citep{simplenet} & \cellcolor{tab_ours}{ViTAD} \\
            \multicolumn{2}{c}{Method~$\rightarrow$} & \makecell[c]{DRAEM$^*$\\~\citep{draem}} & \makecell[c]{UniAD$^\dagger$\\~\citep{uniad}} & \makecell[c]{SimpleNet$^*$\\~\citep{simplenet}} & \cellcolor{tab_ours}{ViTAD} & & \makecell[c]{DRAEM$^*$~\\\cite{draem}} & \makecell[c]{UniAD$^\dagger$\\~\citep{uniad}} & \makecell[c]{SimpleNet$^*$\\~\citep{simplenet}} & \cellcolor{tab_ours}{ViTAD} \\
            \cline{1-2}
            \multicolumn{2}{c}{Category~$\downarrow$} & \red{\textbf{\ding{192}}}~ICCV'21 & \red{\textbf{\ding{194}}}~NeurIPS'22 & \red{\textbf{\ding{193}}}~CVPR'23 & \red{\textbf{\ding{194}}}~\cellcolor{tab_ours}{(Ours)} & & \red{\textbf{\ding{192}}}~ICCV'21 & \red{\textbf{\ding{194}}}~NeurIPS'22 & \red{\textbf{\ding{193}}}~CVPR'23 & \red{\textbf{\ding{194}}}~\cellcolor{tab_ours}{(Ours)} \\
            \hline
            \multirow{4}{*}{\rotatebox{90}{\makecell[c]{Complex \\Structure}}} & PCB1 & 71.9/72.3/70.0 & \underline{94.2}/\underline{92.9}/\underline{90.8} & 91.6/91.9/86.0 & \cellcolor{tab_ours}{\textbf{95.8}{\tiny{\tr{$\pm.22$}}}/\textbf{94.7}{\tiny{\tr{$\pm.43$}}}/\textbf{91.8}{\tiny{\tr{$\pm.95$}}}} & & 94.7/31.9/37.3/52.9 & \underline{99.2}/59.6/59.6/\underline{88.8} & \underline{99.2}/\textbf{86.1}/\textbf{78.8}/83.6 & \cellcolor{tab_ours}{\textbf{99.5}{\tiny{\tr{$\pm.01$}}}/\underline{64.5}{\tiny{\tr{$\pm.89$}}}/\underline{61.7}{\tiny{\tr{$\pm.76$}}}/\textbf{89.6}{\tiny{\tr{$\pm.81$}}}} \\
            & \cellcolor{tab_others}{PCB2} & \cellcolor{tab_others}{78.5/78.3/76.3} & \cellcolor{tab_others}{\underline{91.1}/\underline{91.6}/\textbf{85.1}} & \cellcolor{tab_others}{\textbf{92.4}/\textbf{93.3}/84.5} & \cellcolor{tab_ours}{90.6{\tiny{\tr{$\pm.12$}}}/89.9{\tiny{\tr{$\pm.36$}}}/\textbf{85.3}{\tiny{\tr{$\pm.82$}}}} & & \cellcolor{tab_others}{92.3/\underline{10.0}/\underline{18.6}/66.2} & \cellcolor{tab_others}{\textbf{98.0}/\pzo9.2/16.9/\underline{82.2}} & \cellcolor{tab_others}{96.6/\pzo8.9/\underline{18.6}/\textbf{85.7}} & \cellcolor{tab_ours}{\underline{97.9}{\tiny{\tr{$\pm.06$}}}/\textbf{12.6}{\tiny{\tr{$\pm.40$}}}/\textbf{21.2}{\tiny{\tr{$\pm.49$}}}/82.0{\tiny{\tr{$\pm.53$}}}} \\
            & PCB3 & 76.6/77.5/74.8 & 82.2/83.2/77.5 & \underline{89.1}/\underline{91.1}/\underline{82.6} & \cellcolor{tab_ours}{\textbf{90.9}{\tiny{\tr{$\pm.25$}}}/\textbf{91.2}{\tiny{\tr{$\pm.28$}}}/\textbf{83.9}{\tiny{\tr{$\pm.21$}}}} & & 90.8/14.1/24.4/43.0 & \textbf{98.2}/13.3/24.0/79.3 & \underline{97.2}/\textbf{31.0}/\textbf{36.1}/\underline{85.1} & \cellcolor{tab_ours}{\textbf{98.2}{\tiny{\tr{$\pm.03$}}}/\underline{22.4}{\tiny{\tr{$\pm.87$}}}/\underline{26.4}{\tiny{\tr{$\pm.95$}}}/\textbf{88.0}{\tiny{\tr{$\pm.71$}}}} \\
            & \cellcolor{tab_others}{PCB4} & \cellcolor{tab_others}{97.4/97.6/93.5} & \cellcolor{tab_others}{\underline{99.0}/\textbf{99.1}/\underline{95.5}} & \cellcolor{tab_others}{97.0/97.0/93.5} & \cellcolor{tab_ours}{\textbf{99.1}{\tiny{\tr{$\pm.13$}}}/\underline{98.9}{\tiny{\tr{$\pm.18$}}}/\textbf{96.6}{\tiny{\tr{$\pm.27$}}}} & & \cellcolor{tab_others}{94.4/\underline{31.0}/\underline{37.6}/75.7} & \cellcolor{tab_others}{\underline{97.2}/29.4/33.5/\underline{82.9}} & \cellcolor{tab_others}{93.9/23.9/32.9/61.1} & \cellcolor{tab_ours}{\textbf{99.1}{\tiny{\tr{$\pm.03$}}}/\textbf{42.9}{\tiny{\tr{$\pm.47$}}}/\textbf{48.3}{\tiny{\tr{$\pm.51$}}}/\textbf{91.8}{\tiny{\tr{$\pm.79$}}}} \\
            \hline
            \multirow{4}{*}{\rotatebox{90}{\makecell[c]{Multiple \\Instances}}} & Macaroni1 & 69.8/68.6/70.9 & 82.8/79.3/\underline{75.7} & \textbf{85.9}/\underline{82.5}/73.1 & \cellcolor{tab_ours}{\underline{85.8}{\tiny{\tr{$\pm.40$}}}/\textbf{83.9}{\tiny{\tr{$\pm.74$}}}/\textbf{76.7}{\tiny{\tr{$\pm.57$}}}} & & 95.0/\textbf{19.1}/\textbf{24.1}/67.0 & \textbf{99.0}/\pzo7.6/16.1/\textbf{92.6} & \underline{98.9}/\pzo3.5/\pzo8.4/\underline{92.0} & \cellcolor{tab_ours}{98.5{\tiny{\tr{$\pm.01$}}}/\pzo\underline{8.0}{\tiny{\tr{$\pm.73$}}}/\underline{19.3}{\tiny{\tr{$\pm.88$}}}/89.2{\tiny{\tr{$\pm.47$}}}} \\
            & \cellcolor{tab_others}{Macaroni2} & \cellcolor{tab_others}{59.4/60.7/68.1} & \cellcolor{tab_others}{\underline{76.0}/\textbf{75.8}/\underline{70.2}} & \cellcolor{tab_others}{68.3/54.3/59.7} & \cellcolor{tab_ours}{\textbf{79.1}{\tiny{\tr{$\pm.79$}}}/\underline{74.7}{\tiny{\tr{$\pm.85$}}}/\textbf{74.9}{\tiny{\tr{$\pm.83$}}}} & & \cellcolor{tab_others}{94.6/\pzo\underline{3.9}/\textbf{12.5}/65.3} & \cellcolor{tab_others}{\underline{97.3}/\pzo\textbf{5.1}/\underline{12.2}/\underline{87.0}} & \cellcolor{tab_others}{93.2/\pzo0.6/\pzo3.9/77.8} & \cellcolor{tab_ours}{\textbf{98.1}{\tiny{\tr{$\pm.03$}}}/\pzo3.6{\tiny{\tr{$\pm.02$}}}/10.4{\tiny{\tr{$\pm.24$}}}/\textbf{87.2}{\tiny{\tr{$\pm.03$}}}} \\
            & Capsules & \textbf{83.4}/\textbf{91.1}/\textbf{82.1} & 70.3/83.2/77.8 & 74.1/82.8/74.6 & \cellcolor{tab_ours}{\underline{79.2}{\tiny{\tr{$\pm.62$}}}/\underline{87.6}{\tiny{\tr{$\pm.17$}}}/\underline{79.8}{\tiny{\tr{$\pm.72$}}}} & & 97.1/27.8/33.8/62.9 & \underline{97.4}/\underline{40.4}/\underline{44.7}/72.2 & 97.1/\textbf{52.9}/\textbf{53.3}/\underline{73.7} & \cellcolor{tab_ours}{\textbf{98.2}{\tiny{\tr{$\pm.03$}}}/30.4{\tiny{\tr{$\pm.82$}}}/41.4{\tiny{\tr{$\pm.81$}}}/\textbf{75.1}{\tiny{\tr{$\pm.09$}}}} \\
            & \cellcolor{tab_others}{Candle} & \cellcolor{tab_others}{69.3/73.9/68.1} & \cellcolor{tab_others}{\textbf{95.8}/\textbf{96.2}/\textbf{90.0}} & \cellcolor{tab_others}{84.1/73.3/76.6} & \cellcolor{tab_ours}{\underline{90.4}{\tiny{\tr{$\pm.45$}}}/\underline{91.2}{\tiny{\tr{$\pm.53$}}}/\underline{83.7}{\tiny{\tr{$\pm.91$}}}} & & \cellcolor{tab_others}{82.2/10.1/19.0/65.6} & \cellcolor{tab_others}{\textbf{99.0}/\textbf{23.6}/\textbf{32.6}/\textbf{93.0}} & \cellcolor{tab_others}{\underline{97.6}/\pzo8.4/16.5/\underline{87.6}} & \cellcolor{tab_ours}{96.2{\tiny{\tr{$\pm.09$}}}/\underline{16.8}{\tiny{\tr{$\pm.19$}}}/\underline{26.4}{\tiny{\tr{$\pm.21$}}}/85.2{\tiny{\tr{$\pm.21$}}}} \\
            \hline
            \multirow{4}{*}{\rotatebox{90}{\makecell[c]{Single \\Instance}}} & Cashew & 81.7/89.7/\underline{87.3} & \textbf{94.3}/\textbf{97.2}/\textbf{91.1} & \underline{88.0}/91.3/84.7 & \cellcolor{tab_ours}{87.8{\tiny{\tr{$\pm.83$}}}/\underline{94.2}{\tiny{\tr{$\pm.38$}}}/86.1{\tiny{\tr{$\pm.47$}}}} & & 80.7/\pzo9.9/15.8/38.5 & \textbf{99.0}/56.2/58.9/\textbf{88.5} & \underline{98.9}/\textbf{68.9}/\textbf{66.0}/\underline{84.1} & \cellcolor{tab_ours}{98.5{\tiny{\tr{$\pm.06$}}}/\underline{63.9}{\tiny{\tr{$\pm.38$}}}/\underline{62.7}{\tiny{\tr{$\pm.24$}}}/78.8{\tiny{\tr{$\pm.83$}}}} \\
            & \cellcolor{tab_others}{Chewing Gum} & \cellcolor{tab_others}{93.7/97.2/91.0} & \cellcolor{tab_others}{\textbf{97.5}/\textbf{98.9}/\textbf{96.4}} & \cellcolor{tab_others}{\underline{96.4}/\underline{98.2}/\underline{93.8}} & \cellcolor{tab_ours}{94.9{\tiny{\tr{$\pm.18$}}}/97.7{\tiny{\tr{$\pm.06$}}}/91.4{\tiny{\tr{$\pm.63$}}}} & & \cellcolor{tab_others}{91.1/\textbf{62.4}/\textbf{63.3}/41.0} & \cellcolor{tab_others}{\textbf{99.1}/59.5/58.0/\textbf{85.0}} & \cellcolor{tab_others}{\underline{97.9}/26.8/29.8/\underline{78.3}} & \cellcolor{tab_ours}{97.8{\tiny{\tr{$\pm.04$}}}/\underline{61.6}{\tiny{\tr{$\pm.83$}}}/\underline{58.7}{\tiny{\tr{$\pm.88$}}}/71.5{\tiny{\tr{$\pm.47$}}}} \\
            & Fryum & \underline{89.2}/\underline{95.0}/\underline{86.6} & 86.9/93.9/86.0 & 88.4/93.0/83.3 & \cellcolor{tab_ours}{\textbf{94.3}{\tiny{\tr{$\pm.28$}}}/\textbf{97.4}{\tiny{\tr{$\pm.11$}}}/\textbf{90.9}{\tiny{\tr{$\pm.26$}}}} & & 92.4/38.8/38.6/69.5 & \underline{97.3}/\underline{46.6}/\textbf{52.4}/82.0 & 93.0/39.1/45.4/\underline{85.1} & \cellcolor{tab_ours}{\textbf{97.5}{\tiny{\tr{$\pm.02$}}}/\textbf{47.1}{\tiny{\tr{$\pm.08$}}}/\underline{50.3}{\tiny{\tr{$\pm.21$}}}/\textbf{87.8}{\tiny{\tr{$\pm.76$}}}} \\
            & \cellcolor{tab_others}{Pipe Fryum} & \cellcolor{tab_others}{82.8/91.2/84.0} & \cellcolor{tab_others}{\underline{95.3}/\underline{97.6}/\underline{92.9}} & \cellcolor{tab_others}{90.8/95.5/88.6} & \cellcolor{tab_ours}{\textbf{97.8}{\tiny{\tr{$\pm.11$}}}/\textbf{99.0}{\tiny{\tr{$\pm.05$}}}/\textbf{94.7}{\tiny{\tr{$\pm.37$}}}} & & \cellcolor{tab_others}{91.1/38.2/39.7/61.9} & \cellcolor{tab_others}{\underline{99.1}/53.4/58.6/\underline{93.0}} & \cellcolor{tab_others}{98.5/\underline{65.6}/\underline{63.4}/83.0} & \cellcolor{tab_ours}{\textbf{99.5}{\tiny{\tr{$\pm.02$}}}/\textbf{66.0}{\tiny{\tr{$\pm.92$}}}/\textbf{66.5}{\tiny{\tr{$\pm.52$}}}/\textbf{94.7}{\tiny{\tr{$\pm.34$}}}} \\
            \hline
            \multicolumn{2}{c}{Average} & 79.5/82.8/79.4 & \underline{88.8}/\underline{90.8}/\underline{85.8} & 87.2/87.0/81.7 & \cellcolor{tab_ours}{\textbf{90.5}{\tiny{\tr{$\pm.09$}}}/\textbf{91.7}{\tiny{\tr{$\pm.22$}}}/\textbf{86.3}{\tiny{\tr{$\pm.19$}}}} & & 91.4/24.8/30.4/59.1 & \textbf{98.3}/33.7/\underline{39.0}/\textbf{85.5} & 96.8/\underline{34.7}/37.8/81.4 & \cellcolor{tab_ours}{\underline{98.2}{\tiny{\tr{$\pm.00$}}}/\textbf{36.6}{\tiny{\tr{$\pm.21$}}}/\textbf{41.1}{\tiny{\tr{$\pm.15$}}}/\underline{85.1}{\tiny{\tr{$\pm.25$}}}} \\
            \hline
            \multicolumn{2}{c}{mAD} & - & - & - & - & & 63.9 & \underline{74.5} & 72.4 & \textbf{75.6} \\
            \bottomrule
        \end{tabular}
}
\end{table*}

\begin{table*}[thb]
    \centering
    \caption{\textbf{Multi-class image-level anomaly classification and pixel-level anomaly segmentation results on Uni-Medical.}}
    \label{tab:medical}
    \renewcommand{\arraystretch}{1.2}
    \setlength\tabcolsep{1.0pt}
    \resizebox{1.0\linewidth}{!}{
        % \begin{tabular}{p{2.3cm}<{\centering} p{2.0cm}<{\centering} p{2.0cm}<{\centering} p{2.0cm}<{\centering} p{3.6cm}<{\centering} p{0.1cm}<{\centering} p{2.7cm}<{\centering} p{2.7cm}<{\centering} p{2.7cm}<{\centering} p{4.7cm}<{\centering}}
        \begin{tabular}{cccccccccc}
            \toprule
            & \multicolumn{4}{c}{\textbf{Image-level} mAU-ROC/mAP/m$F_1$-max} & & \multicolumn{4}{c}{\textbf{Pixel-level} mAU-ROC/mAP/m$F_1$-max/mAU-PRO} \\
            \cline{2-5} \cline{7-10}
            % Method~$\rightarrow$ & DRAEM$^*$~\citep{draem} & UniAD$^\dagger$~\citep{uniad} & SimpleNet$^*$~\citep{simplenet} & \cellcolor{tab_ours}{ViTAD} & & DRAEM$^*$~\citep{draem} & UniAD$^\dagger$~\citep{uniad} & SimpleNet$^*$~\citep{simplenet} & \cellcolor{tab_ours}{ViTAD} \\ 
            Method~$\rightarrow$ & \makecell[c]{DRAEM$^*$\\~\citep{draem}} & \makecell[c]{UniAD$^\dagger$\\~\citep{uniad}} & \makecell[c]{SimpleNet$^*$\\~\citep{simplenet}} & \cellcolor{tab_ours}{ViTAD} & & \makecell[c]{DRAEM$^*$\\~\citep{draem}} & \makecell[c]{UniAD$^\dagger$\\~\citep{uniad}} & \makecell[c]{SimpleNet$^*$\\~\citep{simplenet}} & \cellcolor{tab_ours}{ViTAD} \\ 
            \cline{1-1}
            Category~$\downarrow$ & \red{\textbf{\ding{192}}}~ICCV'21 & \red{\textbf{\ding{194}}}~NeurIPS'22 & \red{\textbf{\ding{193}}}~CVPR'23 & \red{\textbf{\ding{194}}}~\cellcolor{tab_ours}{(Ours)} & & \red{\textbf{\ding{192}}}~ICCV'21 & \red{\textbf{\ding{194}}}~NeurIPS'22 & \red{\textbf{\ding{193}}}~CVPR'23 & \red{\textbf{\ding{194}}}~\cellcolor{tab_ours}{(Ours)} \\
            \hline
            Brain & 69.2/90.1/90.7 & \underline{89.9}/\textbf{97.5}/\underline{92.6} & 82.3/\underline{95.6}/90.9 & \cellcolor{tab_ours}{\textbf{90.1}{\tiny{\tr{$\pm.33$}}}/\textbf{97.5}{\tiny{\tr{$\pm.28$}}}/\textbf{93.1}{\tiny{\tr{$\pm.56$}}}} &       & 52.0/\pzo4.8/\pzo6.1/12.5 & \underline{97.4}/\underline{55.7}/\underline{55.7}/\underline{82.4} & 94.8/42.1/42.4/73.0 & \cellcolor{tab_ours}{\textbf{97.8}{\tiny{\tr{$\pm.03$}}}/\textbf{65.3}{\tiny{\tr{$\pm.38$}}}/\textbf{61.8}{\tiny{\tr{$\pm.26$}}}/\textbf{84.0}{\tiny{\tr{$\pm.41$}}}} \\
            \cellcolor{tab_others}{Liver} & 59.1/\underline{52.7}/60.7 & \underline{61.0}/48.8/\underline{63.2} & 55.8/47.6/60.9 & \cellcolor{tab_ours}{\textbf{64.2}{\tiny{\tr{$\pm.63$}}}/\textbf{55.4}{\tiny{\tr{$\pm.72$}}}/\textbf{65.1}{\tiny{\tr{$\pm.49$}}}} &       & 52.9/\pzo1.1/\pzo1.3/\pzo6.6 & 97.1/\pzo7.8/13.7/\textbf{92.7} & \underline{97.4}/\underline{13.2}/\underline{20.1}/86.3 & \cellcolor{tab_ours}{\textbf{98.0}{\tiny{\tr{$\pm.05$}}}/\textbf{14.1}{\tiny{\tr{$\pm.73$}}}/\textbf{22.0}{\tiny{\tr{$\pm.37$}}}/\underline{90.5}{\tiny{\tr{$\pm.53$}}}} \\
            Retinal & 51.7/43.8/59.6 & 84.6/79.4/73.9 & \underline{88.8}/\underline{87.6}/\underline{78.6} & \cellcolor{tab_ours}{\textbf{92.1}{\tiny{\tr{$\pm.28$}}}/\textbf{90.1}{\tiny{\tr{$\pm.38$}}}/\textbf{82.0}{\tiny{\tr{$\pm.27$}}}} &       & 57.4/\pzo6.6/11.3/\pzo0.9 & 94.8/49.3/51.3/79.9 & \underline{95.5}/\underline{59.5}/\underline{56.3}/\underline{82.1} & \cellcolor{tab_ours}{\textbf{95.9}{\tiny{\tr{$\pm.02$}}}/\textbf{70.3}{\tiny{\tr{$\pm.32$}}}/\textbf{65.0}{\tiny{\tr{$\pm.51$}}}/\textbf{83.9}{\tiny{\tr{$\pm.40$}}}} \\
            \hline
            Average & 60.0/62.2/70.3 & \underline{78.5}/75.2/76.6 & 75.6/\underline{76.9}/\textbf{76.8} & \cellcolor{tab_ours}{\textbf{82.2}{\tiny{\tr{$\pm.25$}}}/\textbf{81.0}{\tiny{\tr{$\pm.37$}}}/\textbf{80.1}{\tiny{\tr{$\pm.31$}}}} &       & 54.1/\pzo4.1/\pzo6.2/\pzo6.6 & \underline{96.4}/37.6/\underline{40.2}/\underline{85.0} & 95.9/\underline{38.3}/39.6/80.5 & \cellcolor{tab_ours}{\textbf{97.2}{\tiny{\tr{$\pm.01$}}}/\textbf{49.9}{\tiny{\tr{$\pm.35$}}}/\textbf{49.6}{\tiny{\tr{$\pm.32$}}}/\textbf{86.1}{\tiny{\tr{$\pm.37$}}}} \\
            \hline
            mAD & - & - & - & - & & 37.7 & \underline{69.9} & 69.1 & \textbf{75.2} \\
            \bottomrule
        \end{tabular}
}
\end{table*}

\vspace{1mm}
\noindent\textbf{Evaluation Metrics for AD.} 
Similar to~\citep{rd,draem,visa,uninformedstudents}, we use threshold-independent measures, including 
mean Area Under the Receiver Operating Curve (mAU-ROC) to evaluate binary classification ability, mean Average Precision~\citep{draem} (mAP) to calculate the area under the PR curve, and mean Area Under the Per-Region-Overlap~\citep{uninformedstudents} (mAU-PRO) to weigh regions of different size equally.
In addition, threshold-dependent mean $F_1$-score at optimal threshold~\citep{visa} (m$F_1$-max) is employed to relieve the potential data imbalance problem.
Note that mAU-ROC, mAP, and m$F_1$-max are used in
image-level (anomaly classification) and pixel-level (anomaly segmentation) evaluations. 
The maximum pixel-level value is regarded as the image-level anomaly score~\citep{rd,uniad}.
The models are evaluated ten times evenly for all methods, and the result corresponding to the maximum pixel-level mAU-ROC value is taken as the final result. 
We demonstrate and emphasize using all metrics for evaluation, and the mean of all the metrics termed mAD is reported to indicate the overall performance. 

\vspace{1mm}
\noindent\textbf{Comparison Methods.} 
As MUAD is a relatively new task, we mainly evaluate the published UniAD methods~\citep{uniad}.
We also compare with the latest Augmentation-based DRAEM~\citep{draem}, Reconstruction-based RD~\citep{rd}, and Embedding-based DeSTSeg~\citep{destseg}/SimpleNet~\citep{simplenet}. 
%
%MH: I have no idea what you mean by this. In addition, it is written in Bad English. 
%R: Thanks, this expression can be very ambiguous, modified!
%The official codes extend these methods to fair training for the unified MUAD setting. 
Since the above methods only report results under the SUAD setting, we retrain them to obtain MUAD results by official codes.
In addition, we replace the ViT~\citep{vit} encoder of our ViTAD with various pretrained manners to ensure fair and comprehensive evaluations, including pretrained models by ImageNet~\citep{imagenet} and self-supervised MoCo~\citep{moco}, MAE~\citep{mae}, CLIP~\citep{clip}, DINOv2~\citep{dino2}, and DINO~\citep{dino}.
We use five criteria to compare current concurrent to show differences among them clearly. 
As shown in \cref{tab:model_criterion}, comparison methods more or less require 1) pyramidal encoder, 2) heavy fuser to aggregate multi-depth features, 3) pyramidal decoder, 4) multi-resolution features to keep accurate anomaly location ability, and 5) feature augmentation to obtain better results. 
The design of ViTAD is simple but effective without resorting to elaborate structure or complex augmentations.

\vspace{1mm}
\noindent\textbf{Training.} 
ViTAD chooses ViT-S with DINO-S weights as the structure of the encoder and decoder, and it is trained on images of 256 $\times$ 256 pixels without other data augmentations for all experiments. 
The AdamW optimizer~\citep{adamw} is used with an initial learning rate of 1e$^{-4}$, a weight decay of 1e$^{-4}$, and a batch size of 8. 
Our model only requires 100 training epochs on a single GPU in all experiments, and the learning rate drops by 0.1 after 80 epochs.
All images are trained together without any categorical labels for the MUAD setting, and the weights of the encoder are frozen by default as previous methods~\citep{rd,uniad,simplenet}.
Note that the above hyperparameters are fixed in all experiments under different settings and datasets for our approach without elaborate fine-tuning.

\subsection{Performance Evaluation} \label{exp:sotas}
% \label{exp:mvtec} \label{exp:visa} \label{exp:medical}
\noindent \textbf{Quantitative Evaluations on MVTec AD.} 
We evaluate the ViTAD method with state-of-the-art approaches using both image-level (\cref{tab:mvtec_sp}) and pixel-level (\cref{tab:mvtec_px}) metrics on the MVTec AD dataset. 
The proposed ViTAD method performs favorably against all the evaluated schemes, even when trained solely with the plain columnar ViT and a simple cosine similarity loss. 
ViTAD achieves better image-level results than UniAD with SoTA results on mAU-ROC/mAP/m$F_1$-max of 98.3/99.4/97.3.
In addition, ViTAD achieves performance gain of \rub{0.7}/\rub{10.2}/\rub{8.3}/\rub{0.7} using pixel-level metrics (mAU-ROC/mAP/m$F_1$-max of 97.7/55.3/58.7/91.4) and performance gain of \rub{3.0} using the mean metric (85.4).  

We have a few findings from these empirical results. 
First, models with higher single-class results do not necessarily perform better in multi-class scenarios, as shown in the comparison between UniAD~\citep{uniad} and SimpleNet~\citep{simplenet}. 
This could be due to model over-fitting or single-class-specific training strategies. 
Second, different methods yield similar results for categories under texture but show significant differences for categories with semantic objects, reflecting the diversity and effectiveness of different methods. 
Third, our method performs favorably across all categories, with no categories scoring particularly low (for instance, all mAU-ROC scores are above 90.0), demonstrating the effectiveness and generalizability of our method. In contrast, other methods invariably underperform in specific categories. 
Fourth, even without employing pyramidal encoders and decoders, our method still achieves state-of-the-art anomaly segmentation results, demonstrating the inherent fine-grained multi-scale modeling capability of the plain ViT.

\begin{figure*}[tp]
    \centering
    \includegraphics[width=1.0\linewidth]{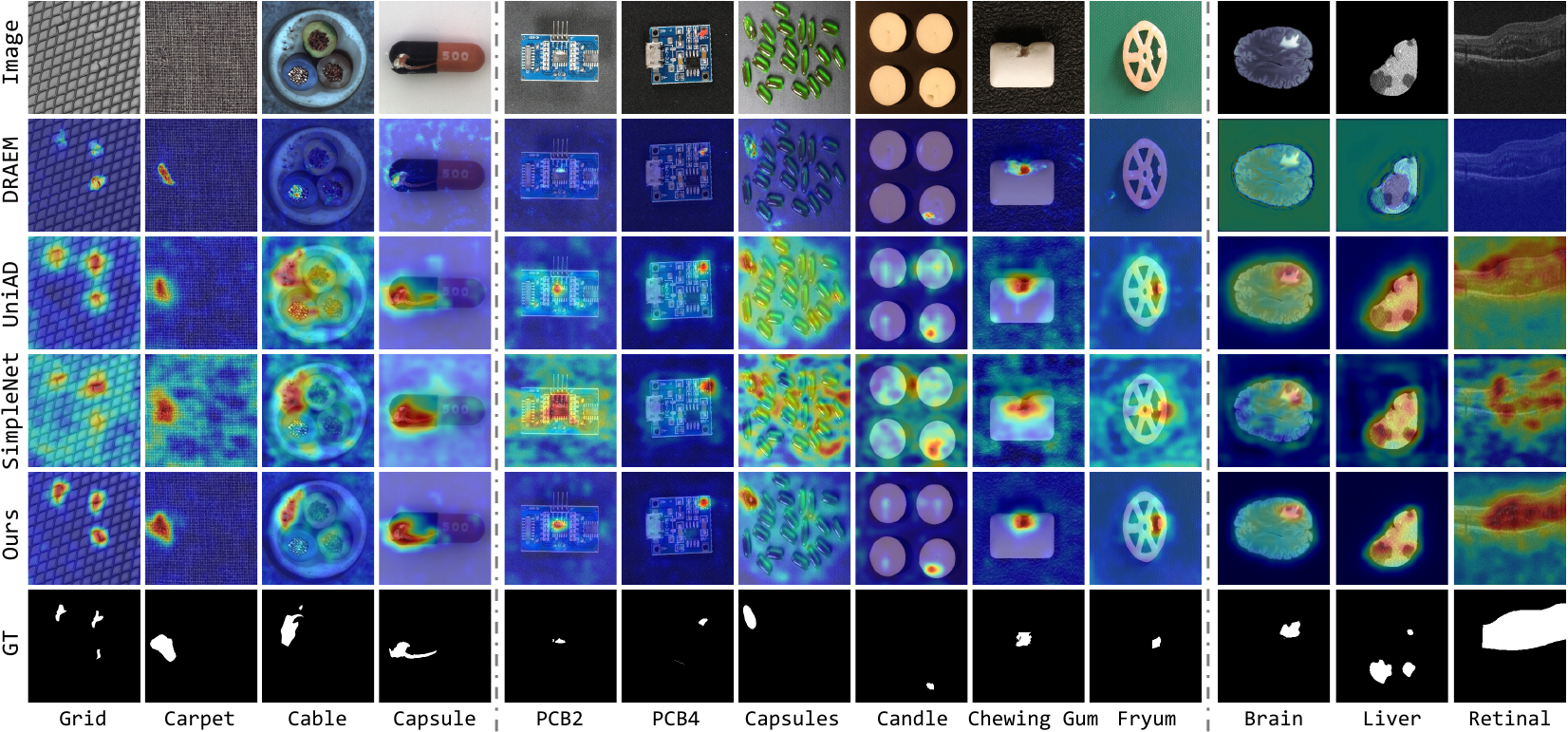}
    \caption{\textbf{Qualitative visualized results for anomaly segmentation}. Compared with the latest augmentation-based DRAEM~\citep{draem} (2nd row), reconstruction-based UniAD~\citep{uniad} (3rd row), and embedding-based SimpleNet~\citep{simplenet} (4th row) on MVTec AD~\citep{mvtec} (\textbf{Left}), VisA~\citep{visa} (\textbf{Middle}), and Uni-Medical~\citep{bmad} (\textbf{Right}) datasets, our ViTAD (5th row) has a more accurate and compact anomaly location capability.}
    \label{fig:vis_comparison}
\end{figure*}

% In fact, ~\citep{draem} is oddly named DRAEM rather than DREAM ... i.e., Discriminatively trained Reconstruction Anomaly Embedding Model. 

\noindent \textbf{Quantitative Evaluations on VisA.} 
The VisA dataset contains more complex structures, multiple and large variations of objects, and more images. 
We use one of the most advanced and powerful methods from each category for comprehensive performance evaluations.
The quantitative results in \cref{tab:visa} show that ViTAD consistently performs well against state-of-the-art schemes. 
ViTAD surpasses UniAD by mAU-ROC/mAP/m$F_1$-max of \rub{1.7}/\rub{0.9}/\rub{0.5}.
In addition, it obtains 75.6 on the comprehensive mAD metric, exceeding UniAD by \rub{1.1}. 

\noindent \textbf{Quantitative Evaluations on Uni-Medical.} 
Compared to industrial AD datasets, Uni-Medical presents more significant challenges due to the more difficult anomaly types and a more comprehensive range of anomaly areas. 
Following the fair training and evaluation setting, \cref{tab:medical} benchmarks quantitative results compared to the state-of-the-art methods. 
ViTAD achieves a significant advantage, reaching 75.2 mAD that surpasses the second-best UniAD by \rub{5.3}, while the performance of DRAEM significantly decreases. 
This indicates that our ViTAD has strong generalization ability across different types of datasets.

\noindent \textbf{Qualitative Evaluations on Three AD Datasets.} 
\label{exp:qualitative}
We conduct qualitative experiments to analyze the anomaly localization performance of evaluated methods. 
\cref{fig:vis_comparison} shows the anomaly detection results on various object categories in MVTec AD~\citep{mvtec}, VisA~\citep{visa}, and Uni-Medical~\citep{bmad} datasets. 
Compared to the augmentation-based DRAEM, reconstruction-based UniAD, and the embedding-based SimpleNet, our method can find more accurate and compact anomalous areas with less edge uncertainty and shows fewer false positives in normal areas. 
Using examples of a textural carpet with large-scale anomalies and a capsule object with irregularly shaped anomalies, ViTAD segments anonymous areas more accurately with fewer false positive responses.

\begin{table}[tp]
    \caption{\textbf{Efficiency comparison of different methods.}}
    \renewcommand{\arraystretch}{1.0}
    \setlength\tabcolsep{2.0pt}
    \resizebox{1.0\linewidth}{!}{
        % \begin{tabular}{p{1.7cm}<{\raggedright} p{1.5cm}<{\centering} p{1.5cm}<{\centering} p{1.8cm}<{\centering} p{1.7cm}<{\centering} p{1.7cm}<{\centering}}
        \begin{tabular}{lcccccc}
        \toprule
        Method & Parameters & FLOPs & Train Memory & Train Time & Train Epoch & FPS \\
        \toprule
        DRAEM & \pzo97.4~M & 198.0~G & 19,852~M & 19.6~H & 700 & \pzo54.0 \\
        RD & \pzo80.6~M & \pzo28.4~G & \uwave{\pzo3,872~M} & \uwave{\pzo4.1~H} & 200 & \underline{\pzo90.6} \\
        UniAD & \textbf{\pzo24.5~M} & \textbf{\pzo\pzo3.6~G} & \pzo6,844~M & 13.4~H & 1,000 & \pzo56.8 \\
        % RD++ & 176.6~M & \underline{\pzo\pzo5.0~G} & \pzo7,794~M & \pzo9.5~H \\
        DeSTSeg & \underline{\pzo35.2~M} & 122.7~G & \underline{\pzo3,562~M} & \underline{\pzo2.5~H} & 660 & \uwave{\pzo74.7} \\
        SimpleNet & \pzo72.8~M & \uwave{\pzo16.1~G} & \pzo5,488~M & 11.8~H & \underline{200} & \pzo49.3 \\
        \rowcolor{tab_ours} ViTAD (Ours) & \uwave{\pzo38.6~M} & \underline{\pzo10.7~G} & \textbf{\pzo2,300~M} & \textbf{\pzo1.1~H} & \textbf{100} & \textbf{112.3} \\ 
        \bottomrule
        \end{tabular}
    }
    \label{tab:ab_efficiency}
\end{table}

\noindent \textbf{Efficiency Comparison.} 
We evaluate the model efficiency in five aspects: 
1) number of parameters, 2) FLOPs, 3) train memory, 4) run-time (evaluated on a V100 GPU with a batch size of 8), and 5) train epoch. 
\cref{tab:ab_efficiency} shows that ViTAD requires significantly smaller memory for training and run-time while achieving SoTA performance. 
It is worth noting that ViTAD only requires 1.1 hours and 2.3G of GPU memory for training while requiring very competitive parameters and FLOPs.

\noindent \textbf{Summary.} 
Results demonstrate that using only non-pyramidal ViT is sufficient to achieve state-of-the-art performance, and our ViTAD can be well generalized to various AD datasets in different domains. 
This indicates that a pyramidal structure for the encoder/decoder is unnecessary for constructing AD models. 

\subsection{Ablation Studies} \label{exp:ablation}

% \lxt{Two shortcomings of ablation studies: 1, it seems they are not well aligned with the method. 2, need more words to explain why this results and the connection to our design. Such as, `` Thus, we keep xxx design in Sec.""}

\subsubsection{Global Structural Designs for ViTAD} \label{exp:global}
\begin{figure}[tp]
    \centering
    \includegraphics[width=1.0\linewidth]{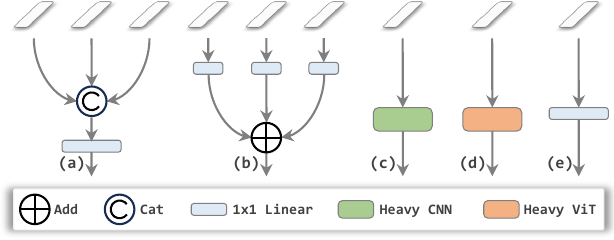}
    \caption{\textbf{Alternative Fuser {\textit{\bm{$\mathcal{F}$}}}}.   
    (a) Only the last feature $F_{4}$.
    (b) Multi-scale features with concatenation
    (c) Multi-scale features with addition; \textbf{(d)} $F_{4}$ followed by a heavy CNN network; \textbf{(e)} $F_{4}$ followed by a heavy ViT network.}
    \label{fig:fuser}
\end{figure}

\noindent\textbf{Structural Variants of Fuser \textit{\bm{$\mathcal{F}$}}.}
In addition to the structure described in \cref{section:vitad}, we analyze different Fuser structures by exploiting two fusion schemes of multi-layer features in \cref{fig:fuser}(a)-(b). 
\cref{tab:ab_fuser} shows corresponding quantitative evaluation results.
Concatenation and addition schemes exhibit negligible result differences. 
In addition, gradually integrating multi-scale features initially decreases and then slightly increases performance, indicating that multi-scale features are not the optimal choice for AD tasks. 
%
%MH: need to explain a bit more if you want to relate these findings to RD and UniAD (why?).
% These findings differ slightly from the reported results in RD~\citep{rd} and UniAD~\citep{uniad} where multi-scale features are shown to be useful for performance gains. 
The phenomenon differs from the structural design of RD~\citep{rd} and UniAD~\citep{uniad}, where multi-scale features are deemed necessary for performance gains.
Considering model complexity and performance, we only use the output feature of the last layer, \ie, $F_4$, as the input to the decoder. 
This strikes a good balance between the accuracy and performance of the AD model. 
Furthermore, we explore the effect of deeper sub-models with CNN-based BottleNeck (\cref{fig:fuser}(c)) and ViT block (\cref{fig:fuser}(d)) as the input to the decoder. 
Additional CNN-based structure can significantly enhance the model performance in all metrics, termed ViTAD-C in \colorbox{oran_tab}{\textcolor{orange}{orange}} background. 
These results indicate that complementary basic structures can bring additional gains for our ViTAD, even though the plain structure has already achieved impressive SoTA results. 
For instance, the average mAD increases from 85.4 to 86.2 (\rub{0.8}) after adding one BottleNeck layer, but more layers do not significantly improve the performance. 
In contrast, using ViT does not bring about a noticeable improvement in model performance. 
Following the principle of simple and effective structural design, ViTAD uses one linear layer as the Fuser structure (\cref{fig:fuser}(e)) and keeps this design in \cref{eq:fuser}. 

\begin{table}[tb]
    \caption{\textbf{Quantitative evaluation of different Fuser \textit{\bm{$\mathcal{F}$}} variants.} The numbers in the ``Cat'' and ``Add'' rows at the top represent the usage of corresponding stage features, while the numbers in the ``ViT'' and ``Conv'' rows at the bottom denote the number of ViT~\citep{vit} and BottleNeck~\citep{resnet} layers used respectively. Methods in \colorbox{lblu_tab}{\textcolor{blue}{blue}} and \colorbox{oran_tab}{\textcolor{orange}{orange}} represent our ViTAD and ViTAD-C, respectively.}
    \renewcommand{\arraystretch}{1.2}
    \setlength\tabcolsep{2pt}
    \resizebox{1.0\linewidth}{!}{
        % \begin{tabular}{p{0.3cm}<{\centering} p{0.8cm}<{\centering} p{1.4cm}<{\centering} p{0.8cm}<{\centering} p{1.4cm}<{\centering} p{0.1cm}<{\centering} p{1.4cm}<{\centering} p{0.8cm}<{\centering} p{1.4cm}<{\centering} p{1.4cm}<{\centering} p{0.1cm}<{\centering} p{0.8cm}<{\centering}}
        \begin{tabular}{cccccccccccc}
        \toprule
        \multicolumn{2}{c}{\multirow{2}{*}{\makecell[c]{Fuser \\Variant}}} & \multicolumn{3}{c}{Image-level} & & \multicolumn{4}{c}{Pixel-level} & & \multirow{2}{*}{mAD} \\
        \cline{3-5} \cline{7-10}
        & & mAU-ROC & mAP & m$F_1$-max & & mAU-ROC & mAP & m$F_1$-max & mAU-PRO & & \\
        \toprule
        \multirow{4}{*}{\rotatebox{90}{\makecell[c]{Cat}}} & 01234 & 97.5 & 99.0 & 96.4 &  & 97.8 & 55.3 & 59.1 & 91.5 &  & 85.2 \\
        & 1234 & 97.6 & 99.0 & 96.4 &  & 97.8 & 55.2 & 59.1 & 91.2 &  & 85.2 \\
        & 234 & 97.4 & 98.8 & 96.3 &  & 97.2 & 54.3 & 57.7 & 91.0 &  & 84.7 \\
        & 34 & 97.8 & 98.9 & 96.8 &  & 97.4 & 54.7 & 58.3 & 91.4 &  & 85.0 \\
        \hline
        \multirow{4}{*}{\rotatebox{90}{\makecell[c]{Add}}} & 01234 & 97.5 & 99.0 & 96.5 &  & 97.8 & 55.4 & 59.3 & 91.3 &  & 85.3 \\
        & 1234 & 97.5 & 99.0 & 96.5 &  & 97.8 & 55.2 & 59.1 & 91.2 &  & 85.2 \\
        & 234 & 97.3 & 98.6 & 96.1 &  & 97.2 & 54.3 & 58.0 & 90.8 &  & 84.6 \\
        & 34 & 98.2 & 99.3 & 97.1 &  & 97.3 & 54.4 & 58.2 & 91.1 &  & 85.1 \\
        \hline
        \multirow{3}{*}{\rotatebox{90}{\makecell[c]{ViT}}} & 1 & 98.2 & 99.3 & 97.2 &  & 97.6 & 55.4 & 58.5 & 91.2 &  & 85.4 \\
        & 3 & 98.4 & 99.4 & 97.3 &  & 97.7 & 55.7 & 58.6 & 91.3 &  & 85.5 \\
        & 5 & 98.3 & 99.3 & 97.5 &  & 97.7 & 55.4 & 58.7 & 91.3 &  & 85.5 \\
        \hline
        \multirow{3}{*}{\rotatebox{90}{\makecell[c]{Conv}}} & \cellcolor{oran_tab}{1} & \cellcolor{oran_tab}{98.4} & \cellcolor{oran_tab}{99.4} & \cellcolor{oran_tab}{97.6} &  & \cellcolor{oran_tab}{98.0} & \cellcolor{oran_tab}{57.6} & \cellcolor{oran_tab}{60.2} & \cellcolor{oran_tab}{91.9} &  & \cellcolor{oran_tab}{86.2} \\
        & 3 & 97.9 & 99.2 & 97.1 &  & 98.0 & 57.8 & 60.7 & 91.9 &  & 86.1 \\
        & 5 & 98.1 & 99.3 & 97.1 &  & 98.1 & 58.5 & 60.7 & 91.8 &  & 86.2 \\
        \hline
        & \cellcolor{tab_ours}{ViTAD} & \cellcolor{tab_ours}{98.3} & \cellcolor{tab_ours}{99.4} & \cellcolor{tab_ours}{97.3} &  & \cellcolor{tab_ours}{97.7} & \cellcolor{tab_ours}{55.3} & \cellcolor{tab_ours}{58.7} & \cellcolor{tab_ours}{91.4} &  & \cellcolor{tab_ours}{85.4} \\
        \bottomrule
        \end{tabular}
    }
    \label{tab:ab_fuser}
\end{table}

\begin{table}[tp]
    \caption{\textbf{Ablation study on model depth and division.} $i \times j$ indicates that the network contains $i$ divisions with each owing $j$ layers for the decoder. $a - b - c$ means that the decoder has 3 divisions with $a$, $b$, and $c$ layers, respectively.}
    \renewcommand{\arraystretch}{1.0}
    \setlength\tabcolsep{6.0pt}
    \resizebox{1.0\linewidth}{!}{
        % \begin{tabular}{p{1.5cm}<{\centering} p{1.7cm}<{\raggedright} p{2.2cm}<{\centering} p{3.0cm}<{\centering} p{1.2cm}<{\centering} }
        \begin{tabular}{clccc}
        \toprule
        Encoder & Decoder & Image-level & Pixel-level & mAD \\
        \toprule
        1 $\times$ 3 & 1 $\times$ 3 & 69.9/83.6/86.4 & 81.1/21.9/27.1/58.4 & 61.2 \\
        2 $\times$ 3 & 1 $\times$ 3 & 93.6/96.9/93.6 & 93.8/48.6/52.3/87.1 & 80.8 \\
        3 $\times$ 3 & 2 $\times$ 3 & 98.1/99.2/97.2 & 97.2/53.9/57.6/91.2 & 84.9 \\
        \hline
        4 $\times$ 3 &  3 $\times$ 2 & 98.0/99.2/97.2 & 97.6/55.0/58.5/91.3 & 85.3 \\
        4 $\times$ 3 &  3 $\times$ 4 & 98.1/99.2/97.2 & 97.7/55.4/58.7/91.2 & 85.3 \\
        4 $\times$ 3 & 2 $-$ 3 $-$ 4 & 98.1/99.1/97.1 & 97.7/55.3/58.7/91.5 & 85.4 \\
        4 $\times$ 3 & 4 $-$ 3 $-$ 2 & 98.1/99.2/97.2 & 97.6/55.1/58.4/91.0 & 85.2 \\
        \hline
        6 $\times$ 2 & 5 $\times$ 2 & 98.1/99.2/97.1 & 97.6/55.5/58.7/91.5 & 85.4 \\
        3 $\times$ 4 & 2 $\times$ 4 & 97.8/99.1/97.3 & 97.9/55.7/59.0/91.4 & 85.4 \\
        \rowcolor{tab_ours} 4 $\times$ 3 & 3 $\times$ 3 & 98.3/99.4/97.3 & 97.7/55.3/58.7/91.4 & 85.4 \\
        \bottomrule
        \end{tabular}
    }
    \label{tab:ab_depth_division}
\end{table}

\noindent\textbf{Model Depth and Division Analysis.} 
Considering the computational cost and performance of the model, a 12-layer ViT-S with three layers for each division is utilized for experiments. 
Only the last three divisions, \aka stages in other contexts, are employed for the decoder. 
In addition to this division pattern, we experiment with different depth and division combinations, as shown in \cref{tab:ab_depth_division}. 
1) The top part demonstrates the effect of different depths of the encoder and decoder on the results, with shallower model depths leading to poorer performance. 
2) The middle part shows the effect of an asymmetric decoder, which is found to have a negligible impact on the results, which also indirectly demonstrates the stability and robustness of our approach. 
3) The bottom part indicates that different division methods have a minimal effect on the results, as these models fully consider both deep and shallow features. 
When using $i$ divisions in the encoder, the output of the last division is fed into the Fuser, while the outputs of other $i-1$ divisions are used as restraint features. 
Since the features of the first stage are not used, the number of divisions in the decoder is always one less than that in the encoder.
Considering both performance and structural generality, we choose the configuration from the last row as the final structure that is illustrated above (\cref{section:vitad_global_local}). 

\begin{table}[tp]
    \caption{\textbf{Empirical study on different pretrained weights of ViT-S.} IN1K: pretrained on ImageNet-1K. IN22K: first pretrained on ImageNet-22K, then finetuned on ImageNet-1K. $^*$: with patch size equaling 8. $^\dagger$: with further 384 resolution. $^{O}$: Open encoder training. Note that ViT-B is used for MAE and CLIP due to the absence of the ViT-S model.}
    \renewcommand{\arraystretch}{1.1}
    \setlength\tabcolsep{2pt}
    \resizebox{1.0\linewidth}{!}{
        % \begin{tabular}{p{1.0cm}<{\raggedright} p{1.4cm}<{\centering} p{0.8cm}<{\centering} p{1.4cm}<{\centering} p{0.1cm}<{\centering} p{1.4cm}<{\centering} p{0.8cm}<{\centering} p{1.4cm}<{\centering} p{1.4cm}<{\centering} p{0.1cm}<{\centering} p{0.8cm}<{\centering}}
        \begin{tabular}{lcccccccccc}
        \toprule
        \multirow{2}{*}{Model} & \multicolumn{3}{c}{Image-level} & & \multicolumn{4}{c}{Pixel-level} & & \multirow{2}{*}{mAD}\\
        \cline{2-4} \cline{6-9}
        & mAU-ROC & mAP & m$F_1$-max & & mAU-ROC & mAP & m$F_1$-max & mAU-PRO & & \\
        \toprule
        Rand & 59.5 & 79.5 & 84.7 &  & 74.7 & 15.3 & 20.4 & 45.7 &  & 54.2 \\
        IN1K$^{O}$ & 68.5 & 84.0 & 86.1 &  & 76.0 & 15.6 & 21.4 & 49.8 &  & 57.3 \\
        \hline
        IN1K & 94.4 & 97.5 & 95.3 &  & 96.6 & 51.6 & 54.5 & 87.2 &  & 82.4 \\
        IN22K & 95.6 & 97.7 & 95.5 &  & 97.1 & 51.6 & 55.3 & 87.7 &  & 82.9 \\
        DeiT & 95.8 & 98.1 & 96.1 &  & 97.1 & 53.9 & 56.8 & 87.8 &  & 83.7 \\
        CLIP & 71.2 & 84.5 & 85.7 &  & 81.6 & 19.4 & 25.1 & 56.8 &  & 60.6 \\
        MoCo & 95.3 & 97.7 & 95.2 &  & 97.4 & 53.0 & 56.2 & 90.6 &  & 83.6 \\
        MAE & 95.3 & 97.7 & 95.2 &  & 97.4 & 53.0 & 56.2 & 90.6 &  & 83.6 \\
        DINOv2 & 93.2 & 96.8 & 94.9 &  & 96.1 & 47.4 & 52.4 & 86.8 &  & 81.1 \\
        \rowcolor{tab_ours} DINO & 98.3 & 99.4 & 97.3 &  & 97.7 & 55.3 & 58.7 & 91.4 &  & 85.4 \\
        \hline
        DINO$^*$ & 97.0 & 98.8 & 96.6 &  & 98.2 & 63.2 & 63.6 & 93.2 &  & 87.2 \\
        DINO$^\dagger$ & 97.3 & 98.9 & 96.3 &  & 98.3 & 66.7 & 65.7 & 93.4 &  & 88.1 \\
        \bottomrule
        \end{tabular}
    }
    \label{tab:ab_pretrain}
\end{table}
% finetune

\noindent\textbf{Effect of pretrained ViT.} 
We analyze the effect of different pretrained ViT models on anomaly detection. 
\cref{tab:ab_pretrain} shows that the pretrained models are crucial for the subsequent anomaly detection. 
1) The model with random weights still performs reasonably (first row), as the image-level and pixel-level mAU-ROC values are higher than 0.5, and image-level mAP and mF1 reach 79.5 and 84.7, respectively 
Furthermore, we use ImageNet-1K pretrained weights to initialize the encoder and open it for training (second row). However, this is much worse than the frozen encoder (\cf, the third row). 
This is because the finetuned weight on the small AD dataset will reduce the original feature distribution, resulting in a weak expression capability of encoder features. While AD approaches highly depend on rich expression as previous works~\citep{rd,uniad,simplenet} acknowledged. 
2) The AD performance depends heavily on the adopted pretrained ViTs as shown in Rows 3 to 10 of \cref{section:setup}. 
For example, the mAD of DINO is \rub{3.0} higher than the supervised training model of ImageNet-1K, and \rub{1.8} higher than the self-supervised training MAE. 
The results based on CLIP are significantly worse than those of other methods, in which the training objective is to align images and texts that disregard information for detailed structures. 
DINOv2~\citep{dino2} does not perform as well as DINO on the AD task, demonstrating once again the significance of the pretrained weights.
Moreover, the performance of the pretrained model in AD does not correlate with its corresponding classification accuracy. 
For instance, the classification accuracy of DINO-Small, MoCo v3-Small, and MAE-Small on ImageNet-1K are 82.8, 83.2, and 83.6~\citep{mae}, respectively, but on the contrary, DINO-Small significantly better than other evaluated  approaches~\cref{tab:ab_pretrain}. 
3) Based on DINO pretrained weights, we further explore the effect of smaller patch size and higher image resolution on the results (\cf, rows 11 and row 12). 
Both manners would increase computation costs. 
Nevertheless, they slightly reduce the image-level indicators but significantly increase the pixel-level results, \eg, mAU-ROC, mAP, m$F_1$-max, and mAU-PRO increase by up to \rub{0.6}, \rub{11.4}, \rub{7.0}, and \rub{2.0}, and the averaged mAD increases from 85.4 to 88.1. 
Considering both performance and computation, we utilize DINO~\citep{dino} as the pre-training model for the encoder.
% \noindent\textbf{Pyramidal backbone architecture and pre-trained weights.} 
4) Furthermore, we investigate the impact of pyramidal backbone architecture and pre-trained weights on comparative methods. As shown in \cref{tab:backbone}, using the same reconstruction-based RD as an example, the model accuracy decreases when the pre-trained weights are switched to DINO. This indicates that the method is highly sensitive to pre-trained weights. In contrast, our method significantly benefits from DINO weights. Additionally, when the backbone is replaced with a pyramidal Swin Transforme~\citep{swin}, the performance also declines. This demonstrates that RD cannot leverage the ViT structure as effectively as our ViTAD, further proving the superiority of our architecture. 

\begin{table}[tp]
    \caption{\textbf{Research on pyramidal backbone architecture and pre-trained weights for reconstruction-based RD.}}
    \renewcommand{\arraystretch}{1.0}
    \setlength\tabcolsep{6.0pt}
    \resizebox{1.0\linewidth}{!}{
        % \begin{tabular}{p{1.5cm}<{\centering} p{1.7cm}<{\raggedright} p{2.2cm}<{\centering} p{3.0cm}<{\centering} p{1.2cm}<{\centering} }
        \begin{tabular}{ccccc}
        \toprule
        Method & Pretrain & Image-level & Pixel-level & mAD \\
        \toprule
        RD~\citep{rd} & IN1K & 94.6/96.5/95.2 & 96.1/48.6/\underline{53.8}/\underline{91.2} & 82.3 \\
        RD~\citep{rd} & DINO & 89.0/94.7/92.7 & 94.8/47.4/51.3/89.2 & 79.9 \\ 
        RD-Swin~\citep{swin} & IN1K & 90.2/94.9/92.9 & 95.1/47.6/52/88.6 & 80.2 \\
        \rowcolor{tab_ours} ViTAD & DINO & 98.3/99.4/97.3 & 97.7/55.3/58.7/91.4 & 85.4 \\
        \bottomrule
        \end{tabular}
    }
    \label{tab:backbone}
\end{table}

\begin{table}[tp]
    \caption{\textbf{Empirical study on local designs.} \textbf{Before Norm}: $F_4$ is obtained before normalization. \textbf{Add Linear}: Add the linear layer in Fuser. \textbf{Remove CLS Token}: Removing the class token throughout the procedure. \textbf{Use Pos. Embed.}: Keeping the position embedding in the decoder.}
    \renewcommand{\arraystretch}{1.0}
    \setlength\tabcolsep{3.0pt}
    \resizebox{1.0\linewidth}{!}{
        % \begin{tabular}{p{1.0cm}<{\centering} p{1.0cm}<{\centering} p{1.5cm}<{\centering} p{1.7cm}<{\centering} p{2.0cm}<{\centering} p{2.3cm}<{\centering} p{1.0cm}<{\centering} }
        \begin{tabular}{ccccccc}
        \toprule
        \textbf{\multirow{2}{*}{\makecell[c]{Before \\ Norm}}} & \textbf{\multirow{2}{*}{\makecell[c]{Add \\ Linear}}} & \textbf{\multirow{2}{*}{\makecell[c]{Remove \\ CLS Token}}} & \textbf{\multirow{2}{*}{\makecell[c]{Use \\ Pos. Embed.}}} & \multirow{2}{*}{\makecell[c]{Image-level}} & \multirow{2}{*}{\makecell[c]{Pixel-level}} & \multirow{2}{*}{\makecell[c]{mAD}} \\
        & & & & & & \\
        \toprule
        \xmarkg & \xmarkg & \xmarkg & \xmarkg & 97.6/99.0/96.8 & 97.5/55.0/58.2/91.0 & 85.0 \\
        \hline
        \cmark & \xmarkg & \xmarkg & \xmarkg & 97.9/99.1/97.0 & 97.5/54.8/58.1/91.1 & 85.1\\
        \xmarkg & \cmark & \xmarkg & \xmarkg & 97.8/99.1/96.8 & 97.6/55.1/58.3/91.2 & 85.1 \\
        \xmarkg & \xmarkg & \cmark & \xmarkg & 97.7/99.1/96.8 & 97.6/55.0/58.3/91.2 & 85.1 \\
        \xmarkg & \xmarkg & \xmarkg & \cmark & 97.6/99.1/96.8 & 97.6/55.2/58.2/91.2 & 85.1 \\
        \hline
        \xmarkg & \cmark & \cmark & \cmark & 98.1/99.2/97.0 & 97.7/55.2/58.5/91.0 & 85.2 \\
        \cmark & \xmarkg & \cmark & \cmark & 98.1/99.2/97.0 & 97.6/55.0/58.3/91.1 & 85.2 \\
        \cmark & \cmark & \xmarkg & \cmark & 98.0/99.1/97.2 & 97.6/55.1/58.3/91.3 & 85.2 \\
        \cmark & \cmark & \cmark & \xmarkg & 98.0/99.1/97.2 & 97.7/55.3/58.5/91.3 & 85.3 \\
        \hline
        \rowcolor{tab_ours} \cmark & \cmark & \cmark & \cmark & 98.3/99.4/97.3 & 97.7/55.3/58.7/91.4 & 85.4 \\
        \bottomrule
        \end{tabular}
    }
    \label{tab:ab_details}
\end{table}

\subsubsection{Local Structural Designs for ViTAD} \label{exp:local}
Based on the improved global designs, we further investigate four subtle structural designs to enhance the performance, \ie, 
i) whether the output for the encoder goes through the final batch normalization. 
ii) whether the feature fuser uses linear feature transformation. 
iii) whether the class token is inherited.  
iv) whether the position embeddings are used in the decoder. 
As shown in \cref{tab:ab_details}, each local design slightly improves the metric results, and the model achieves the best performance when using all components together, \eg, mAD reaches 85.4 that obtains a 0.4 improvement compared to the baseline model. 
Our final ViTAD model takes the configuration of the last row.

\begin{table}[tp]
    \caption{\textbf{Ablation study on model scaling.} Besides DINO-Small/Base models, results of ViT trained on ImageNet-1K at different scales are also presented. Pretrained models are from tripartite TIMM of version v0.8.15dev0.}
    \renewcommand{\arraystretch}{1.1}
    \setlength\tabcolsep{2pt}
    \resizebox{1.0\linewidth}{!}{
        % \begin{tabular}{p{0.2cm}<{\raggedright} p{0.8cm}<{\centering} p{1.4cm}<{\centering} p{0.8cm}<{\centering} p{1.4cm}<{\centering} p{0.1cm}<{\centering} p{1.4cm}<{\centering} p{0.8cm}<{\centering} p{1.4cm}<{\centering} p{1.4cm}<{\centering} p{0.1cm}<{\centering} p{0.8cm}<{\centering}}
        \begin{tabular}{lccccccccccc}
        \toprule
        \multicolumn{2}{c}{\multirow{2}{*}{\makecell[c]{Backbone \\Scale}}} & \multicolumn{3}{c}{Image-level} & & \multicolumn{4}{c}{Pixel-level} & & \multirow{2}{*}{mAD} \\
        \cline{3-5} \cline{7-10}
        & & mAU-ROC & mAP & m$F_1$-max & & mAU-ROC & mAP & m$F_1$-max & mAU-PRO & & \\
        \toprule
        \multirow{5}{*}{\rotatebox{90}{\makecell[c]{ImageNet-1K}}} & T & 89.7 & 95.2 & 93.0 &  & 94.3 & 43.8 & 49.7 & 82.7 &  & 78.3 \\
        & S & 94.4 & 97.5 & 95.3 &  & 96.6 & 51.6 & 54.5 & 87.2 &  & 82.4 \\
        & B & 96.1 & 98.1 & 96.5 &  & 96.0 & 50.7 & 54.6 & 86.8 &  & 82.7 \\
        & L & 94.9 & 97.6 & 94.3 &  & 94.6 & 45.2 & 49.6 & 85.7 &  & 80.3 \\
        & H & 90.9 & 95.6 & 92.2 &  & 93.8 & 45.0 & 48.3 & 79.8 &  & 77.9 \\
        \midrule
        \multirow{2}{*}{\rotatebox{90}{\makecell[c]{DINO}}} & S & 98.3 & 99.4 & 97.3 &  & 97.7 & 55.3 & 58.7 & 91.4 &  & 85.4 \\
        & B & 97.5 & 99.0 & 96.7 &  & 97.9 & 55.4 & 58.6 & 90.1 &  & 85.0 \\
        \bottomrule
        \end{tabular}
    }
    \label{tab:ab_scaling}
\end{table}

\subsubsection{Model Scale Analysis} 
\label{exp:ab_scaling}
We analyze the ViTAD scale on anomaly detection based on two pretrained settings: supervised learning with ImageNet-1K~\citep{imagenet} and unsupervised learning with DINO features~\citep{dino}. 
As illustrated in \cref{tab:ab_scaling}, larger-scale models do not continually improve AD performance, contrary to recent findings in the classification, detection, and segmentation tasks.
For instance, the image-level mAU-ROC achieved by the DINO-B model is 0.8 lower than that by the DINO-S model.
In addition, the mAD achieved by the DINO-B model is 0.4 lower than that by the DINO-S model.
Thus, we choose ViT-S with DINO-S weights as the detailed structure.  
%This phenomenon is also reflected in AD works such as RD~\citep{rd}, UniAD~\citep{uniad}, \etc 
%
% These findings are similar to the report results in RD~\citep{rd} and UniAD~\citep{uniad}.
%
%MH: in the paper? what paper?
%For example, when switching to a larger ResNet-152~\citep{resnet} model or a more powerful Swin Transformer~\citep{swin} backbone, the results are significantly inferior to those of the default backbone in the paper. 
%
% For example, when switching to a larger ResNet-152~\citep{resnet} model or a more powerful Swin Transformer~\citep{swin} backbone, the results are significantly worse than those of the original model in [CITE A PAPER].
%
%This inspires us to explore the inconsistency between the trend of model accuracy growth and larger-scale models in future research.

\begin{figure}[tp]
    \centering
    \includegraphics[width=1.0\linewidth]{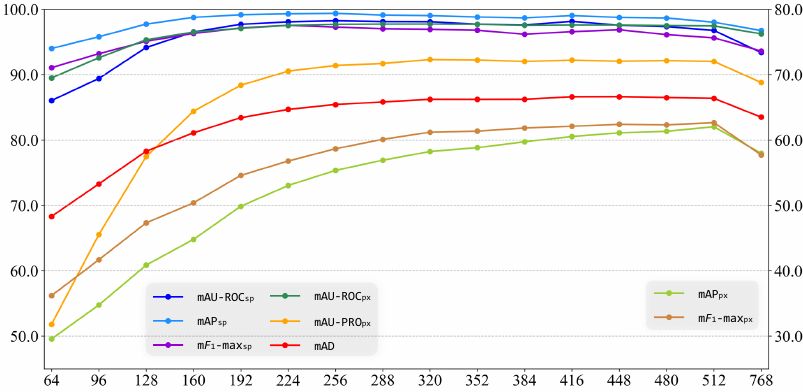}
    \caption{\textbf{Model performance of ViTAD with varying resolutions.} Pixel-level mAP$_{px}$ and m$F_1$-max$_{px}$ use the right vertical axis, while the remaining metrics share the left vertical axis.}
    \label{fig:resolution}
\end{figure}

\begin{table}[tb]
    \caption{\textbf{Empirical study on restraint stages.} Numbers represent the constrained stages.}
    \renewcommand{\arraystretch}{1.1}
    \setlength\tabcolsep{2pt}
    \resizebox{1.0\linewidth}{!}{
        % \begin{tabular}{p{1.36cm}<{\centering} p{1.4cm}<{\centering} p{0.8cm}<{\centering} p{1.4cm}<{\centering} p{0.1cm}<{\centering} p{1.4cm}<{\centering} p{0.8cm}<{\centering} p{1.4cm}<{\centering} p{1.4cm}<{\centering} p{0.1cm}<{\centering} p{0.8cm}<{\centering}}
        \begin{tabular}{ccccccccccc}
        \toprule
        \multirow{2}{*}{\makecell[c]{Restraint \\ Stage}} & \multicolumn{3}{c}{Image-level} & & \multicolumn{4}{c}{Pixel-level} & & \multirow{2}{*}{mAD}\\
        \cline{2-4} \cline{6-9}
        & mAU-ROC & mAP & m$F_1$-max & & mAU-ROC & mAP & m$F_1$-max & mAU-PRO & & \\
        \toprule
        $A_{3}$ & 95.2 & 98.0 & 94.6 &  & 96.9 & 50.0 & 54.7 & 87.1 &  & 82.4 \\
        $A_{2}$,$A_{3}$ & 97.0 & 98.8 & 96.3 &  & 97.4 & 52.3 & 57.1 & 89.2 &  & 84.0 \\
        \rowcolor{tab_ours} $A_{1}$,$A_{2}$,$A_{3}$ & 98.3 & 99.4 & 97.3 &  & 97.7 & 55.3 & 58.7 & 91.4 &  & 85.4 \\
        $A_{0}$,$A_{1}$,$A_{2}$,$A_{3}$ & 95.5 & 97.6 & 95.2 &  & 97.4 & 53.5 & 57.4 & 90.5 &  & 83.9 \\
        \bottomrule
        \end{tabular}
    }
    \label{tab:ab_restraint}
\end{table}

\begin{table}[t!]
    \caption{\textbf{Ablation study on different pixel-wise loss functions of different approaches.} $^\dagger$: Default loss function in the paper.}
    \renewcommand{\arraystretch}{1.1}
    \setlength\tabcolsep{2pt}
    \resizebox{1.0\linewidth}{!}{
        % \begin{tabular}{p{0.3cm}<{\raggedright} p{0.8cm}<{\raggedright} p{1.4cm}<{\centering} p{0.8cm}<{\centering} p{1.4cm}<{\centering} p{0.1cm}<{\centering} p{1.4cm}<{\centering} p{0.8cm}<{\centering} p{1.4cm}<{\centering} p{1.4cm}<{\centering} p{0.1cm}<{\centering} p{0.8cm}<{\centering}}
        \begin{tabular}{llcccccccccc}
        \toprule
        \multicolumn{2}{c}{\multirow{2}{*}{\makecell[c]{Item}}} & \multicolumn{3}{c}{Image-level} & & \multicolumn{4}{c}{Pixel-level} & & \multirow{2}{*}{mAD} \\
        \cline{3-5} \cline{7-10}
        & & mAU-ROC & mAP & m$F_1$-max & & mAU-ROC & mAP & m$F_1$-max & mAU-PRO & & \\
        \toprule
        \multirow{4}{*}{\rotatebox{90}{\makecell[c]{RD}}} & L1 & 93.4 & 97.2 & 95.7 &  & 95.8 & 46.8 & 52.2 & 90.2 &  & 81.6 \\
        & MSE & 97.7 & 99.0 & 96.7 &  & 96.4 & 48.4 & 53.3 & 91.1 &  & 83.2 \\
        & Cos$_p$ & 95.3 & 97.6 & 96.1 &  & 96.2 & 50.5 & 55.0 & 91.5 &  & 83.2 \\
        & Cos$_f$~$^\dagger$ & 94.6 & 96.5 & 95.2 &  & 96.1 & 48.6 & 53.8 & 91.2 &  & 82.3 \\
        \hline
        \multirow{4}{*}{\rotatebox{90}{\makecell[c]{UniAD}}} & L1 & 96.6 & 98.7 & 96.5 &  & 96.8 & 44.4 & 49.7 & 90.2 &  & 81.8 \\
        & MSE~$^\dagger$ & 97.5 & 99.1 & 97.3 &  & 97.0 & 45.1 & 50.4 & 90.7 &  & 82.4 \\
        & Cos$_p$ & 74.6 & 88.1 & 87.9 &  & 84.1 & 19.1 & 24.6 & 64.2 &  & 63.2 \\
        & Cos$_f$ & 76.5 & 89.1 & 88.5 &  & 82.1 & 18.4 & 24.2 & 62.1 &  & 63.0 \\
        \hline
        \multirow{4}{*}{\rotatebox{90}{\makecell[c]{ViTAD}}} & L1 & 97.6 & 99.0 & 96.8 &  & 97.5 & 54.9 & 58.4 & 91.2 &  & 85.1 \\
        & MSE & 97.7 & 99.1 & 97.0 &  & 97.6 & 55.0 & 58.6 & 91.5 &  & 85.2 \\
        & Cos$_p$ & 98.1 & 99.3 & 97.2 &  & 97.7 & 55.4 & 58.9 & 91.8 &  & 85.5 \\
        & Cos$_f$~$^\dagger$ & 98.3 & 99.4 & 97.3 &  & 97.7 & 55.3 & 58.7 & 91.4 &  & 85.4 \\
        \bottomrule
        \end{tabular}
    }
    \label{tab:ab_loss}
\end{table}

\subsubsection{Resolution Effect} 
\label{exp:ab_resolution}
As existing anomaly detection methods are designed to handle images of 224$\times$224~\citep{rd,simplenet} or 256$\times$256~\citep{draem,uniad,destseg} pixels, there is no analysis of the effect of frame resolution on model performance. 
Considering the practical application requirements for different resolutions, we conduct experiments at intervals of 32 pixels within the range from 64$\times$64 to 512$\times$512 and additionally test the results under images of 768$\times$768 pixels.
%
%As shown in \cref{fig:resolution}, different metric results gradually improve as the resolution increases, and the results tend to stabilize at a resolution of 256$\times$256. 
As shown \cref{fig:resolution}, the performance of ViTAD increases with larger resolution up to 256$\times$256 pixels. 
We note that ViTAD still achieves satisfactory results under a low resolution (\eg, 64$\times$64), but it does not perform well when the resolution is high (\eg, 768$\times$768).
%
%MH: add this line. check the resolution 
As the typical resolution for AD is 256$\times$256 pixels, we leave this for future work. 
%This inspires us to research high-resolution AD tasks in the future.

\subsubsection{Restraint Anomaly Map} \label{exp:ab_restraint_stages}
Based on the standard 4-stage division, \cref{tab:ab_restraint} displays performance under different constrained anomaly maps during the training phase, with all corresponding constrained features computing the anomaly maps. 
The model performs worst when only the last $A_{3}$ is used, while it achieves the best results when the features of the last three anomaly maps are constrained. 
This indicates that the shallow anomaly map ($A_{0}$) would interfere with deep features ($A_{i}, i>0$) that reduces the performance. 
Thus, our ViTAD employs $A_{1}$, $A_{2}$, and $A_{3}$ for the training constraint. 

\begin{table}[tb]
    \caption{\textbf{Ablation study on training epoch, scheduler (Sche.), and augmentation.} CC: Center Crop. CJ: Color Jitter. RHF: Random Horizontal Flip. RR: Random Rotation. RRC: Random Resized Crop.}
    \renewcommand{\arraystretch}{1.1}
    \setlength\tabcolsep{2pt}
    \resizebox{1.0\linewidth}{!}{
        % \begin{tabular}{p{0.3cm}<{\raggedright} p{0.8cm}<{\raggedright} p{1.4cm}<{\centering} p{0.8cm}<{\centering} p{1.4cm}<{\centering} p{0.1cm}<{\centering} p{1.4cm}<{\centering} p{0.8cm}<{\centering} p{1.4cm}<{\centering} p{1.4cm}<{\centering} p{0.1cm}<{\centering} p{0.8cm}<{\centering}}
        \begin{tabular}{llcccccccccc}
        \toprule
        \multicolumn{2}{c}{\multirow{2}{*}{\makecell[c]{Item}}} & \multicolumn{3}{c}{Image-level} & & \multicolumn{4}{c}{Pixel-level} & & \multirow{2}{*}{mAD} \\
        \cline{3-5} \cline{7-10}
        & & mAU-ROC & mAP & m$F_1$-max & & mAU-ROC & mAP & m$F_1$-max & mAU-PRO & & \\
        \toprule
        \multirow{5}{*}{\rotatebox{90}{\makecell[c]{Epoch}}} & 30 & 97.4 & 99.0 & 96.6 &  & 97.5 & 55.2 & 58.7 & 91.1 &  & 85.1 \\
        & 50 & 97.9 & 99.1 & 97.1 &  & 97.7 & 55.5 & 58.8 & 91.3 &  & 85.3 \\
        & \cellcolor{tab_ours}{100} & \cellcolor{tab_ours}{98.3} & \cellcolor{tab_ours}{99.4} & \cellcolor{tab_ours}{97.3} & \cellcolor{tab_ours}{\bluetab{1}} & \cellcolor{tab_ours}{97.7} & \cellcolor{tab_ours}{55.3} & \cellcolor{tab_ours}{58.7} & \cellcolor{tab_ours}{91.4} & \cellcolor{tab_ours}{\bluetab{1}} & \cellcolor{tab_ours}{85.4} \\
        & 200 & 98.1 & 99.1 & 96.7 &  & 97.6 & 55.6 & 58.4 & 91.3 &  & 85.3 \\
        & 300 & 98.1 & 99.1 & 96.9 &  & 97.6 & 55.4 & 58.6 & 91.2 &  & 85.3 \\
        \hline
        \multirow{2}{*}{\rotatebox{90}{\makecell[c]{Sche.}}} & Cosine & 98.0 & 99.0 & 97.1 &  & 97.7 & 55.5 & 58.7 & 91.2 &  & 85.3 \\
        & \cellcolor{tab_ours}{Step} & \cellcolor{tab_ours}{98.3} & \cellcolor{tab_ours}{99.4} & \cellcolor{tab_ours}{97.3} & \cellcolor{tab_ours}{\bluetab{1}} & \cellcolor{tab_ours}{97.7} & \cellcolor{tab_ours}{55.3} & \cellcolor{tab_ours}{58.7} & \cellcolor{tab_ours}{91.4} & \cellcolor{tab_ours}{\bluetab{1}} & \cellcolor{tab_ours}{85.4} \\
        \hline
        \multirow{6}{*}{\rotatebox{90}{\makecell[c]{Augmentation}}} & CC+CJ & 73.1 & 89.8 & 86.3 &  & 66.8 & 8.5 & 15.0 & 20.9 &  & 51.5 \\
        & CC+RHF & 75.6 & 90.9 & 86.7 &  & 65.8 & 8.6 & 14.8 & 19.1 &  & 51.6 \\
        & CC+RR & 72.1 & 88.9 & 85.7 &  & 67.0 & 8.6 & 15.1 & 21.0 &  & 51.2 \\
        & RRC & 92.5 & 96.5 & 92.1 &  & 84.1 & 18.6 & 27.9 & 48.6 &  & 65.8 \\
        & All & 88.3 & 95.0 & 90.3 &  & 75.9 & 13.2 & 21.7 & 35.9 &  & 60.0 \\
        & \cellcolor{tab_ours}{None} & \cellcolor{tab_ours}{98.3} & \cellcolor{tab_ours}{99.4} & \cellcolor{tab_ours}{97.3} & \cellcolor{tab_ours}{\bluetab{1}} & \cellcolor{tab_ours}{97.7} & \cellcolor{tab_ours}{55.3} & \cellcolor{tab_ours}{58.7} & \cellcolor{tab_ours}{91.4} & \cellcolor{tab_ours}{\bluetab{1}} & \cellcolor{tab_ours}{85.4} \\
        \bottomrule
        \end{tabular}
    }
    \label{tab:ab_epoch_scheduler_aug}
\end{table}

\subsubsection{Robustness Evaluation} \label{exp:ab_robustness}
We evaluate the robustness of ViTAD in several aspects. 

\vspace{1mm}
\noindent\textbf{Loss Function.} Following Occam's Razor principle, the reconstruction-based ViTAD only uses a pixel-wise loss function for model training. 
Specifically, we use four types of loss functions, \ie, L1, Mean Square Error (MSE), pixel-wise Cosine Similarity (Cos$_p$), and flattened Cosine Similarity (Cos$_f$), for quantitative comparison with the SoTA RD~\citep{rd} and UniAD~\citep{uniad}. 
As shown in \cref{tab:ab_loss}, our method is robust to different loss functions, with mAD showing no significant fluctuations (bottom part). 
RD has noticeable gaps in multiple metrics (middle part), and UniAD has significant gaps (top part) where a significant decrease in model performance when switching to any Cos$_p$ and Cos$_f$. 

\vspace{1mm}
\noindent\textbf{Train Epoch.} The convergence of the model significantly impacts its application value. 
The top part of \cref{tab:ab_epoch_scheduler_aug} shows the performance under different epochs. 
The proposed method achieves stable results at 30 epochs and optimal results at 100 epochs. 
Considering the balance between training resources and performance, we set the default train epoch to 100, and this number still has a significant advantage compared to the comparison methods (see \cref{section:setup}). 

\vspace{1mm}
\noindent\textbf{Train Scheduler.} The Cosine scheduler has been shown to have a positive effect in the fields of classification, detection, and segmentation; it has not been exploited for anomaly detection. 
%
%Therefore, we explore its impact on our ViTAD in the middle part of \cref{tab:ab_epoch_scheduler_aug}. 
%
\cref{tab:ab_epoch_scheduler_aug} shows the robustness of our approach to different train schedulers, and the effect of the Cosine scheduler slightly decreases compared to the step scheduler. 

\vspace{1mm}
\noindent\textbf{Train Augmentation.} \cref{tab:ab_epoch_scheduler_aug} shows the effect of 5 types of data augmentation on model training.  
We find that proven effective data augmentation methods in other fields may have a negative effect in the AD field. 
This is because the aim of AD is to fit the domain of the training set as closely as possible, and these augmentations could broaden the domain range, resulting in poor test outcomes. 
As the scale of the AD dataset undergoes a substantial transformation, \ie, a much larger AD dataset is proposed, these augmentations might potentially enhance the robustness of the model.
We leave this interesting finding for future work.

\begin{figure}[tp]
    \centering
    \includegraphics[width=1.0\linewidth]{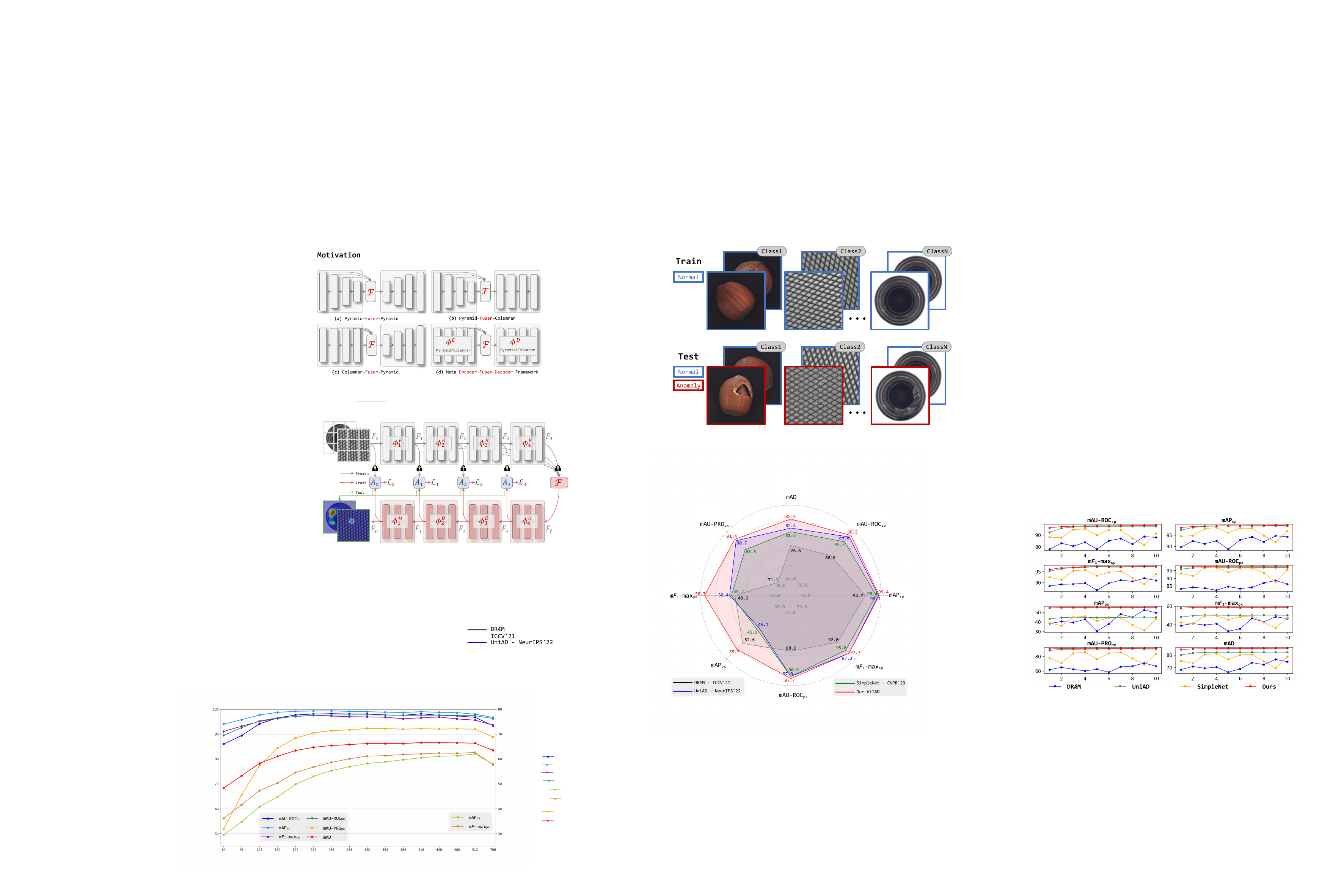}
    \caption{\textbf{Stability comparison of all metrics during the training process for different methods.} Each model is tested ten times at linear intervals during the training process.}
    \label{fig:stable}
\end{figure}

\vspace{1mm}
\noindent\textbf{Metric Stability During Training.} When replicating comparison methods, we find that some methods have significant fluctuations in metric evaluation during training. 
Therefore, we analyze the fluctuations in metrics during training in several mainstream methods. 
As shown in \cref{fig:stable}, DRAEM~\citep{draem} and SimpleNet~\citep{simplenet} have noticeable jitters during training, while UniAD~\citep{uniad} and our method are very stable, but our method has significant advantages in metric results and convergence speed.

\begin{figure}[tp]
    \centering
    \includegraphics[width=1.0\linewidth]{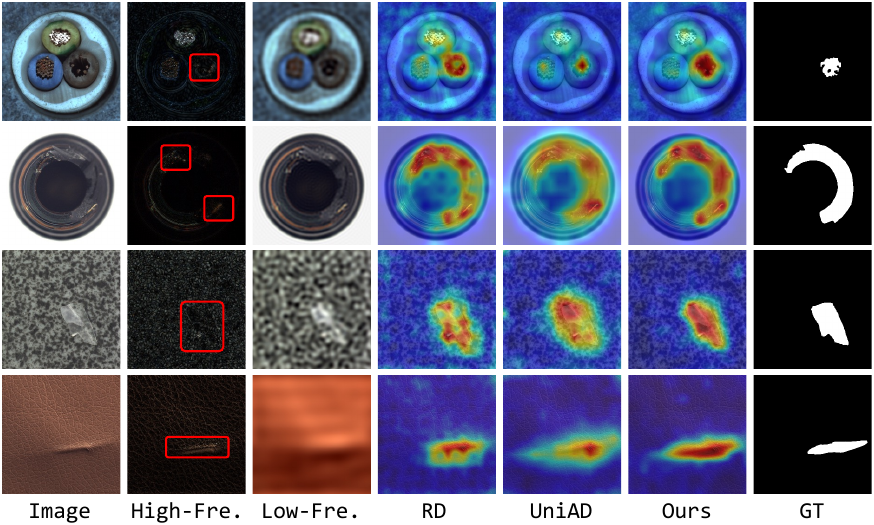}
    \caption{\textbf{Comparative analysis from a frequency domain perspective}. The second and third columns represent the high- and low-frequency decomposition of the input image in the first column, while the last column represents the ground truth of the anomaly segmentation. The other columns display the anomaly segmentation results of different methods.}
    \label{fig:frequency}
\end{figure}

\subsubsection{Advantage Explanation of ViT for AD} \label{exp:ab_frequency}
As shown in \cref{fig:vis_comparison}, ViTAD exhibits a more compact overlap segmentation with the ground truth than the competing methods. 
We attempt to explain the advantage of ViT for the AD task from a frequency perspective. 
Given the substantial distribution difference between abnormal and normal regions, high-frequency information can serve as a representation of the abnormal area. 
Thus, we show the high- and low-frequency components of the original image (second and third columns) in \cref{fig:frequency}, with red rectangular boxes indicating the high-frequency abnormal representation regions. 
ViTAD outperforms comparison methods in segmenting more accurate and compact abnormal regions. 
This is attributed to the global multi-head self-attention module of ViT, which can capture long-range context information and high-frequency details as previously demonstrated~\citep{eatformer,iformer}, while methods that employ a pyramidal structure with low-receptive fields lack this ability.

\section{Conclusion} \label{section:conclusion}
This paper addresses the multi-class unsupervised anomaly detection task using a plain vision transformer. 
Specifically, we abstract a Meta-AD framework based on the current reconstruction methods, and by Occam's Razor principle, we propose a powerful yet efficient ViTAD baseline. 
We propose a comprehensive and fair evaluation benchmark on eight metrics for this increasingly popular task. 
ViTAD achieves impressive results on MVTec AD, VisA, and Uni-Medical datasets without inventing or introducing additional modules, datasets, or training techniques. 
In addition, we conduct thorough experiments to demonstrate the effectiveness and robustness of our method. 
%
%We hope our study will inspire future research for advancing ViT to the MUAD field and beyond. 
%
%\vspace{1mm}
%\noindent\textbf{Limitations and Future Works.} 
%This work only explores the most naive plain ViT solution for MUAD, and further investigations of AD-oriented ViT improvements and data augmentations could potentially enhance the model's performance. 
%
%Additionally, AD-specific pre-training techniques are also worth investigating.
%
%Since ViT can be effortlessly adapted for multi-modal inputs, our model can be generalized to diverse AD applications, \ie, multi-modal 3D AD~\citep{m3dm,easynet} and zero-shot AD~\citep{winclip,aprilgan,clipad}.

\vspace{1mm}
\noindent
\textbf{Broad Impact.} 
This work thoroughly benchmarks mainstream and latest AD methods on three datasets under the challenging MUAD setting. 
Simultaneously, a novel ViTAD is proposed to explore the potential of plain ViT in AD tasks, filling the research gap and stimulating subsequent research works. 
In addition, the proposed ViTAD achieves state-of-the-art results on multiple datasets, with significantly smaller training costs, faster inference speed, and more memory-friendly, demonstrating its greater application values.

% \vspace{1mm}
% \noindent
% \textbf{Acknowledgement.} 
% This work is supported by a Grant from The National Natural Science Foundation of China (No. 62103363).

% \section*{CRediT authorship contribution statement.} 
% \textbf{Jiangning Zhang:} Conceptualization, Data curation, Software, Methodology, Writing – original draft, Writing – review \& editing. 
% \textbf{Xuhai Chen:} Methodology, Visualization, Writing – original draft, Writing – review \& editing. 
% \textbf{Yabiao Wang:} Investigation, Funding acquisition, Writing – review \& editing. 
% \textbf{Chengjie Wang:} Investigation, Funding acquisition, Writing – review \& editing. 
% \textbf{Yong Liu:} Supervision, Methodology, Project administration, Resources, Writing – review \& editing. 
% \textbf{Xiangtai Li:} Methodology, Validation, Writing – review \& editing. 
% \textbf{Ming-Hsuan Yang:} Methodology, Resources, Writing – review \& editing. 
% \textbf{Dacheng Tao:} Methodology, Resources, Writing – review \& editing. 
% % Conceptualization; Data curation; Formal analysis; Funding acquisition; Investigation; Methodology; Project administration; Resources; Software; Supervision; Validation; Visualization; Roles/Writing - original draft; and Writing - review & editing. 

% \section*{Declaration of competing interest.}
% The authors declare that they have no known competing financial interests or personal relationships that could have appeared to influence the work reported in this paper. 

% \section*{Data availability.}
% The dataset used in the manuscript is publicly available.

\bibliographystyle{model2-names}
\bibliography{main}

\end{document}